\documentclass[sigconf]{aamas}

\usepackage{balance} % for balancing columns on the final page
\usepackage{algorithmic}
\usepackage[linesnumbered,ruled,vlined]{algorithm2e}
\usepackage{amsmath}
\usepackage{amsfonts}
\usepackage{multirow}
\usepackage{multicol}

\setcopyright{ifaamas}
\acmConference[AAMAS '26]{Proc.\@ of the 25th International Conference
on Autonomous Agents and Multiagent Systems (AAMAS 2026)}{May 25 -- 29, 2026}
{Paphos, Cyprus}{C.~Amato, L.~Dennis, V.~Mascardi, J.~Thangarajah (eds.)}
\copyrightyear{2026}
\acmYear{2026}
\acmDOI{}
\acmPrice{}
\acmISBN{}

\acmSubmissionID{<<submission id>>}

\title[Active Evaluation]{Active Evaluation of General Agents: Problem\\ Definition and Comparison of Baseline Algorithms}

\author{Marc Lanctot}
\affiliation{
  \institution{Google DeepMind}
  \city{}
  \country{}
  }
\email{lanctot@google.com}

\author{Kate Larson}
\affiliation{
  \institution{Google DeepMind, University of Waterloo}
  \city{}
  \country{}
  }
\email{katelarson@google.com}

\author{Ian Gemp}
\affiliation{
  \institution{Google DeepMind}
  \city{}
  \country{}
  }
\email{imgemp@google.com}

\author{Michael Kaisers}
\affiliation{
  \institution{Google DeepMind}
  \city{}
  \country{}
  }
\email{mkaisers@google.com}

\begin{abstract}
As intelligent agents become more generally-capable, \ie able to master a wide variety of tasks, the complexity and cost of properly evaluating them rises significantly.
Tasks that assess specific capabilities of the agents can be correlated and stochastic, requiring many samples for accurate comparisons, leading to added costs.
In this paper, we propose a formal definition and a conceptual framework for {\it active evaluation} of agents across multiple tasks, which assesses the performance of ranking algorithms as a function of number of evaluation data samples. 
Rather than curating, filtering, or compressing existing data sets as a preprocessing step, we propose an online framing: on every iteration, the ranking algorithm chooses the task and agents to sample scores from. Then, evaluation algorithms report a ranking of agents on each iteration and their performance is assessed with respect to the ground truth ranking over time. Several baselines are compared under different experimental contexts, with synthetic generated data and simulated online access to real evaluation data from Atari game-playing agents. We find that the classical Elo rating system-- while it suffers from well-known failure modes, in theory-- is a consistently reliable choice for efficient reduction of ranking error in practice.
A recently-proposed method, Soft Condorcet Optimization, shows comparable performance to Elo on synthetic data and significantly outperforms Elo on real Atari agent evaluation.
When task variation from the ground truth is high, selecting tasks based on proportional representation leads to higher rate of ranking error reduction.
\end{abstract}


\newcommand{\BibTeX}{\rm B\kern-.05em{\sc i\kern-.025em b}\kern-.08em\TeX}

\newcommand{\bx}{\mathbf{x}}
\newcommand{\by}{\mathbf{y}}

\newcommand{\E}{\mathbb{E}}
\newcommand{\cA}{\mathcal{A}}
\newcommand{\cD}{\mathcal{D}}

\newcommand{\cN}{\mathcal{N}}

\newcommand{\defword}[1]{\textbf{\boldmath{#1}}}
\newcommand{\btheta}{\boldsymbol\theta}
\newcommand{\ie}{{\it i.e.},~}
\newcommand{\eg}{{\it e.g.},~}
\DeclareMathOperator*{\argmax}{arg\,max}

\definecolor{darkgreen}{RGB}{0,125,0}
\definecolor{darkblue}{RGB}{0,0,125}

\newcounter{mlNoteCounter}

\begin{document}

%%% The following commands remove the headers in your paper. For final 
%%% papers, these will be inserted during the pagination process.

\pagestyle{fancy}
\fancyhead{}

%%% The next command prints the information defined in the preamble.

\maketitle 

%%%%%%%%%%%%%%%%%%%%%%%%%%%%%%%%%%%%%%%%%%%%%%%%%%%%%%%%%%%%%%%%%%%%%%%%

\section{Introduction}

Recent progress in AI research has led to intelligent agents which are increasingly more generally-capable. 
This has led to an increase in the number of 
multi-task benchmarks that assess different capabilities or axes of the agents. 
One classic multitask benchmark is the Arcade Learning Environment (ALE), which drove progress on general agent development, leading to the rise of deep reinforcement learning~\cite{bellemare13arcade,Mnih2015DQN}.
The aim was not simply to assess agent performance on a single classic Atari video game, but across a suite of more than fifty different games, similar to general game-playing challenges~\cite{GGP,GVGAI}. 
In recent years, large language models (LLMs) are being more widely adopted by the general public for a variety of different applications. LLMs are being
evaluated for their many different capabilities: mathematical and programming ability, question-answering, general knowledge, abstract reasoning, etc.; however, reporting a complete evaluation of them on a public multi-task benchmark, Holistic Evaluation of Language Models (HELM), can cost thousands of dollars~\cite{liang2022holistic}.
In addition, tasks are often correlated causing redundancy in evaluation data and wasted computational resources. 
Scores attributed to agents/models can also  also be noisy further increasing costs.

In addition to mounting costs, proper multi-task evaluation of general agent abilities is inherently challenging. There are many subtleties involved such as misconceptions in interpretation of results, overfitting to the domain (or evaluation context), and biases~\cite{machado18arcade,singh2025leaderboardillusion,burnell2023rethink}, statistical uncertainty~\cite{rlliable}, trade-offs between diversity and sensitivity~\cite{Zhang24Inherent}, and non-transitivities as well as classical metrics that can be misleading or not robust~\cite{Balduzzi18ReEval,liu2025reevaluatingopenendedevaluationlarge,lanctot2023evaluating}.
While some of these issues are solvable with long-term commitments to better evaluation practices, some are possible to address with guarantees provided by evaluation methodologies rooted in social choice theory and game theory.
For example, recent principled evaluation techniques have offer greater interpretability and robustness based on consistency and invariance properties that are inherited from their game-theoretic and social choice theoretic foundations~\cite{Balduzzi18ReEval,liu2025reevaluatingopenendedevaluationlarge,lanctot2023evaluating,Lanctot25SCO}.

In this paper, we make the following contributions:

\begin{itemize}
\item We formally define the problem of {\it active evaluation} of general agents (\ie those evaluated on multi-task benchmarks); the goal is not only to rank agents but to do so {\it efficiently}. 
\item We define metrics that naturally balance between {\it identifying} the best agents and {\it ranking} them in the correct order.
\item We extend recent offline algorithms for multitask evaluation (Voting-as-Evaluation~\cite{lanctot2023evaluating}, Nash averaging~\cite{Balduzzi18ReEval}, and Metritocracy~\cite{procaccia2025metritocracy}) to the (online) active evaluation setting.
\item We compare our extended algorithms and several baselines on synthetically-generated evaluation data and on an online simulation of Atari agent game-playing data.
\end{itemize}
Unlike traditional multitask evaluation where benchmarks and datasets are procured separately from evaluation, in active evaluation, algorithms are in full control of the data they get to see and choose the tasks and models to score {\it online}.

\section{Background and Terminology}
\label{sec:background}

In this section, we describe the basic terminology that will be used throughout the paper.

A multi-task evaluation problem for general agents consists of:
\begin{itemize}
\item A set of $n$ \defword{tasks} (or environments), $V = \{ v_1, v_2, \cdots, v_n \}$, and 
\item A set of $m$ \defword{agents} (or models), $A = \{ a_1, a_2, \cdots, a_m\}$.
\item A mechanism for assessing the performance of any agent $a_i \in A$ on task $v_k \in V$.
\end{itemize}
A common example in the field of deep reinforcement learning is the Arcade Learning Environment (ALE)~\cite{bellemare13arcade}. In the ALE, $n = 57$ and each task is one of the classic Atari 2600 games;
the agents vary across papers but are commonly deep reinforcement learning algorithms, human agents, and/or reference agents such as random (for example, $m = 8$ in~\cite{Hessel17Rainbow}); the mechanism to assess the performance of an agent is  its average in-game score on the Atari games after thousands of episodes (game plays).
Similarly, multi-task benchmarks have been used for language models~\cite{liang2022holistic} and LLMs as agents~\cite{liu2023agentbench} ($n = 8, m = 27$).

The ultimate goal is to identify the ``best'' agent (or agents) or, most generally, to find a full ranking of agents. This is done by comparing their performance and aggregating them across tasks.
A \defword{ranking} of agents is a total order (permutation) over $A$. For example if $m = 4$, then one potential ranking, $\succ$, denoted
\begin{equation}
a_3 \succ a_0 \succ a_2 \succ a_1, 
\label{eq:example-ranking}
\end{equation}
corresponds to $a_3$ being the top-ranked (best) agent, followed by $a_0$ (second best), $a_2$ (third best), and $a_1$ (worst).
To quantify distances between rankings, such as the number of pairwise disagreements in order of agents, we use Kendall-tau distance.

\begin{definition}
Let $A_1, A_2$ be finite sets of elements such that $A_1 \subseteq A_2$.
Let $\succ_1, \succ_2$ be rankings of agents in $A_1$ and $A_2$, respectively.
The \defword{Kendall-tau distance} between two rankings is defined as 
\begin{equation}
K_d(\succ_1, \succ_2) = \sum_{\{a_i, a_j\} \in A_1, i < j} \textsc{PairwiseOrder}_{i,j}(\succ_1, \succ_2),
\end{equation}
where
$\textsc{PairwiseOrder}_{i,j}(\succ_1, \succ_2) = 0$ if the relative order of $a_i$ and $a_j$ agree across the rankings, \ie either ($a_i \succ_1 a_j$ and $a_i \succ_2 a_j$) or ($a_j \succ_1 a_i$ and $a_j \succ_2 a_i$), or 1 otherwise (\ie they disagree).
\end{definition}

Note that we use a slightly more general definition than the standard Kendall-tau distance (where $A_2$ can be a larger set) to allow computing distances between partial rankings (over $A_1 \subset A_2$) from the full ranking of agents $A_2 = A$.
Since the maximum distance is ${|A_1| \choose 2} = \frac{|A_1| (|A_1| - 1)}{2}$, $K_d$ can normalized to be in $[0,1]$:

\begin{definition}
Let $A_1, A_2$ be finite sets of elements such that $A_1 \subseteq A_2$.
Let $\succ_1, \succ_2$ be permutations over elements in $A_1$ and $A_2$, respectively.
The \defword{normalized Kendall-tau distance} $\succ_1$ and $\succ_2$ is defined as
\begin{equation}
K_n(\pi_1, \pi_2) = \frac{2 K_d(\succ_1, \succ_2)}{|A_1| (|A_1| - 1)}.
\end{equation}
\end{definition}

We will use $K_d$ and $K_n$ ranking distances to the ground truth ranking as a measure of error.

\subsection{Evaluation Systems for General Agents}
\label{sec:eval-systems}

Elo is a classic rating system that uses a simple logistic model learned from win/loss/draw outcomes~\citep{Elo78}. 
A rating, $\theta_i$, is assigned to each player $i$ such that the probability of player $i$ beating player $j$ is predicted as \[
\hat{p}_{ij} = \frac{1}{1 + 10^{(\theta_j - \theta_i)/400}}.
\]
While Elo was designed specifically to rate players in the two-player zero-sum, perfect information game of Chess.
Elo does not inherently model task information and its ratings are computed directly from head-to-head game outcomes.  It is widely used including for the evaluation of LLMs in LMSYS Chatbot Arena~\citep{zheng2023judging}.

Another way to evaluate general agents is to use social choice theory; two previous works are Vote N'Rank or Voting-as-Evaluation (VasE) ~\cite{rofin-etal-2023-votenrank,lanctot2023evaluating}. In VasE, the tasks are {\it votes} and there are no scores (necessarily); the main benefits are axiomatic properties and robustness via various forms of consistency based on the particular voting rule used.
For example, when applying VasE to Atari game-playing, each task is an Atari game (57 tasks in total) and the average score of each agent in the game is sorted to get a complete vote (ranking over agents).
Most recently, there have also been online mechanisms to compute rankings with various properties, such as Soft Condorcet Optimization (SCO) ~\cite{Lanctot25SCO} and KemenyEl~\cite{George24KemenyEl}.
For example, SCO is based on gradient descent of a smooth Kendall-tau distance and the optimum of the SCO's loss function is guaranteed to top-rank a Condorcet winner, if it exists.

Finally, there are game-theoretic methods that can also be used for general agent evaluation, such as Nash Averaging~\cite{Balduzzi18ReEval,Marris25Deviation,Liu25Reevaluating}. These methods derive ratings from equilibria of agents played in an evaluation meta-game; they offer interpretability and robusteness via being invariant to {\bf clones}: informally, a cloned agent is one that mimics the preferences of another agent, not changing any of the relative rankings between any other agents, across all tasks. 

\subsection{Active Learning}

Active learning is a subfield of machine learning (also sometimes called {\it experimental design}) that addresses the challenge of expensive or time-consuming data labeling~\citep[Section 19.4]{pml1Book}. Unlike traditional supervised learning, where a model passively trains on a large, fully labeled dataset, an active learning algorithm is ``curious'': it starts by training on a small number of labeled examples and then intelligently queries a domain expert (or "oracle") to label new data points that it identifies as being the most informative. 

One approach to active learning is based on the information-theoretic principle of information gain that compares the entropy of the classifier on select data points to the expected entropy. This leads to the classical method maximum entropy sampling~\citep{Shewry87}. Maximum entropy sampling is part of a wider class of uncertainty sampling~\citep{Tong02,Settles09} methods that sample preferentially on data points where the classifier is least certain. 
It was shown to be a good baseline approach for active logistic regression~\citep{yang2018benchmark} which could be used as a basis for an active version of Elo.

\subsection{Multi-armed Bandits}
\label{sec:bg-bandits}

In the \defword{stochastic multi-armed bandit} problem, an agent must choose one of several ``arms'' (or options) in each round, with the goal of maximizing the total reward over time. The core assumption is that the reward for pulling any given arm is drawn from a fixed, but unknown, probability distribution. 
The main challenge becomes the exploration-exploitation tradeoff: the agent must {\it explore} by trying different arms to learn their reward distributions, while also {\it exploiting} the arm that currently estimated best to maximize immediate gains. Performance is measured by regret: the cumulative difference between the agent's reward and the reward it would have gotten by always picking the single best arm.
A widely-used stochastic bandit algorithm is Upper Confidence Bounds (UCB)~\citep{Auer02UCB}.

In the \defword{adversarial bandit} (also called non-stochastic) setting~\citep{Bubeck12Bandits}, the problem becomes much harder as all statistical assumptions are dropped. Instead of rewards coming from a fixed distribution, they are chosen by an ``adversary'' in each round. This adversary can set the rewards for every arm arbitrarily, and in the strongest model, can even do so after observing the algorithm's strategy.
Regret is accumulated similarly, but comparing to the best single arm in hindsight. 
Two commonly used algorithms for adversarial bandits are Exp3~\citep{Auer98Exp3} and regret-matching~\citep{Hart01}.

There are well-known connections between two-player zero-sum games and adversarial bandits~\cite{Zinkevich03,PLG,Blum07}: the average strategies of two adversarial bandits in self-play converge to a Nash equilibrium with high probability. This fact will be used to introduce online variants of one VasE algorithm and Nash averaging in Section~\ref{sec:online-ml-nashavg}.

\section{Active Evaluation}
\label{sec:active-eval}

Active evaluation aims to minimize a metric of ranking efficiency. This section discusses several quality metrics for online ranking algorithms, and proposes Average Generalized Ranking Error (AGRE, definition~\ref{def:agre}) as the objective that measures efficiency.

At first the designer must choose an \defword{experimental context}: a \defword{data generator} and arguments for its parameters. 
The experimental context is the main component that ultimately determines the shape of the data.
In this paper, we propose two models for synthetic data generation (see Section~\ref{sec:synthetic-data}) and incremental access to existing data sets.
A task, which may be stochastic, generates {\bf scores} $s(v, a) \sim S_{v,a}$, where $S_{v,a}$ is a random variable that depends on the task and the agent.
Data takes the form of pairwise evaluations $(s(v, a_i), s(v, a_j))$ between agents $a_i, a_j \in A$ on a task $v \in V$.

The main experiment loop is summarized in Algorithm~\ref{alg:active-eval-loop}.
There are $T$ iterations. In each round $t \in \{1, \cdots, T\}$, an algorithm chooses a task $v^t$ and two agents $a_i^t, a_j^t$ and receives data point $s^t$.
The algorithm then uses $s^t$ and outputs a new ranking $\succ^t$. A metric of performance $f$ is used to assess the
performance of the ranking $\succ^t$; we describe several in the next subsection.

\begin{algorithm}[t]
\DontPrintSemicolon % Some LaTeX compilers require you to use \dontprintsemicolon instead
\KwIn{A data generator $\cD$}
\KwIn{A performance metric $f$}
\KwIn{An active evaluation algorithm $\cA$}
\For{round $t \in \{1, 2, \cdots, T\}$}{
   $(v^t, a^t_i, a^t_j) \sim \textsc{ChooseTaskAndAgents}(\cA, A, V, t)$ \label{alg:selection} \;
   Sample scores $s^t = (s(v^t, a^t_i), s(v^t, a_j)) \sim \cD$ \label{alg:sample-scores} \; 
   Update ranking of agents: $\succ^t = \cA(s^t, t)$ \label{alg:update} \;
   Report performance $f(\succ^t)$ at round $t$ \;
}
\Return{final performance $f(\succ^T)$}
\caption{Active Evaluation main loop}
\label{alg:active-eval-loop}
\end{algorithm}

\subsection{Metrics}
\label{sec:metrics}

Many metrics can be used to assess the performance of an algorithm for active evaluation.
Let us first assume the existence of a correct underlying \defword{ground truth ranking}, $\succ^*$ (which is available when generating data synthetically).
In this case, the simplest metric might be to track the Kendall-tau distance to the ground truth over time, $K_d(\succ^t, \succ^*)$. To make the value less dependent on the number of agents, we could use normalized Kendall-tau distance, $K_n$.

However, it is common that the evaluator may only be interested in the top $k$ agents (maybe first, second, and third, \eg $k = 3$) or even more extremely only the top-performing agent ($k = 1$). In these cases, the error metric should only take into account the performance of agents within these top ranks. 

Denote by $\succ_{[k]}$ the sub-ranking of $\succ$ including only the top $k$ agents, and $A_{\succ_{[k]}}$ the set of agents among the top-$k$ ranked agents. For example, in the ranking depicted in equation~(\ref{eq:example-ranking}), $\succ_{[2]} = a_3 \succ a_0$ and $A_{\succ_{[2]}} = \{ a_0, a_3 \}$.

\begin{definition}[Top-$k$ Identification Error]
Let $A$ be a set of agents, $\succ^*$ be a ground truth ranking of agents in $A$ with $|A| = m$, and $k \in \{ 1, \cdots, m \}$.
The \defword{top-$k$ identification error} is the proportion of incorrectly identified top-$k$ agents:
\[
\textsc{IDE}(A, \succ, \succ^*, k) = 1 - \frac{|A_{\succ_{[k]}} \cap A_{\succ^*_{[k]}}|}{k}.
\]
\end{definition}
Since the cardinality of the intersection varies from 0 to $k$, $0 \le \textsc{IDE}(A, \succ, \succ^*, k) \le 1$.
Denote the {\it sub-ranking} $\succ_{k^*}$ as the relative order of the top-$k$ agents in $\succ^*$ within $\succ$.
Now we generalize the ranking error to include both ranking error and top-$k$ identification error:
\begin{definition}[Generalized Top-$k$ Ranking Error]
Let $A$ be a set of agents and $\succ^*$ be a ground truth ranking of agents in $A$, and $k \in \{1, 2, \cdots, m\}$.
The generalized top-$k$ ranking error is a combination of an error for identifying the set of top-$k$ agents and determining their correct rank $\textsc{GRE}(A, \succ, \succ^*, k) = $
\[
\alpha(k) \cdot \textsc{IDE}(A, \succ, \succ^*, k) + (1-\alpha(k)) \cdot K_n(\succ_{k^*}, \succ^*),
\]
where
\[
\alpha(k) = \frac{m - k}{m - 1} \in [0, 1]
\]
is the weight placed on the top-$k$ identification error.
\label{def:gre}
\end{definition}

Since $\alpha(k) + (1 - \alpha(k)) = 1$, and both $\textsc{IDE}(A, \succ, \succ^*, k)$ and $K_n$ are in $[0, 1]$, it follows that $\textsc{GRE}(A, \succ, \succ^*, k) \in [0, 1]$. 
When $k = 1$, all of the weight is placed on identification whereas when $k = m$ all of the weight is placed on Kendall-tau distance, with linear weighting between the two error sources for other values of $k$. This linear weighting naturally places weight proportionally to favor identification error when $k$ is small and Kendall-tau ranking error when $k$ is large.
Finally, we can also think of the {\it average} ranking error similarly to regret analysis of online learning~\citep{PLG,Blum07,HazanOCObook}:

\begin{definition}[Average Generalized Top-$k$ Ranking Error]
Let $A$ be a set of agents, $\succ^*$ be a ground truth ranking of agents in $A$, and $k \in \{1, 2, \cdots, m\}$.
The {\it average generalized ranking error} (AGRE) of an algorithm that outputs rankings $\succ^t$ on round $t$ is:
\[
\textsc{AGRE}(\succ^{1:T}, \succ^*, T) = \frac{\sum_{t=1}^T \textsc{GRE}(A, \succ^t, \succ^*, k)}{T}.
\]
\label{def:agre}
\end{definition}
At any horizon $T$, the AGRE provides a metric of how efficiently the ranking algorithm identifies good rankings, as it cumulates all errors at horizons $t < T$.

\subsection{Models for Synthetic Data}
\label{sec:synthetic-data}

We now describe models for synthetic data generation. 
A key feature of using synthetic data is that the ground truth is known, hence the metrics described in Section~\ref{sec:metrics} can be used directly.

\begin{figure}[t]
    \centering
    \includegraphics[width=1.0\linewidth]{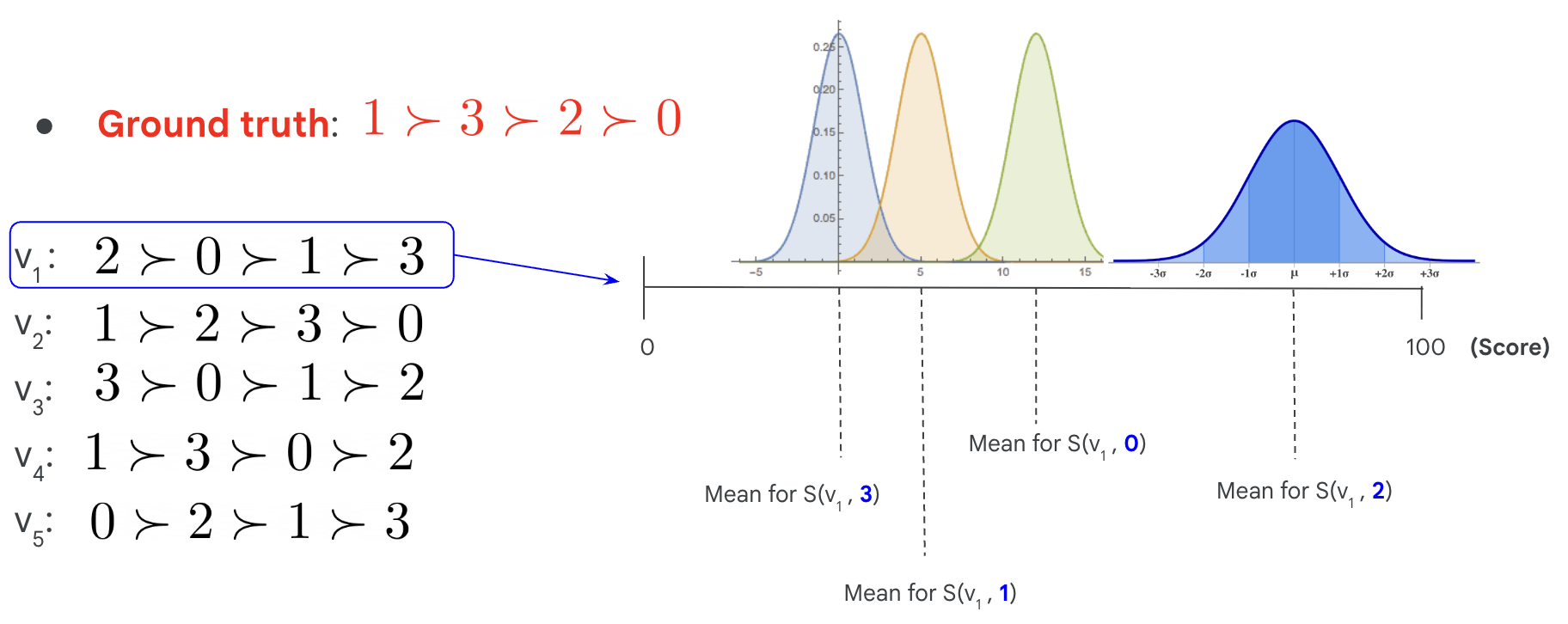}
    \caption{Example of synthetic data generation process with $m = 4$ and $n = 5$. The task rankings $( \succ^*_{v_1}, \succ^*_{v_2}, \cdots, \succ^*_{v_5})$  have Kendall-tau distances 4, 1, 4, 1, and 5, respectively, from $\succ^*$.}
    \label{fig:synthetic-data-gen}
\end{figure}

\subsubsection{Mallows}
\label{sec:mallows}

The Mallows model is a common way to model a number of noisy rankings from a central ranking, $\succ^*$ over $m$ alternatives~\cite{Mallows57,Brandt16Handbook}.
Given a \emph{dispersion} parameter $\phi\in[0,1]$, when $\phi=0$ only the central ranking is sampled; when $\phi=1$ all rankings are equally likely.
The probability of sampling a ranking $\succ$ is
\begin{equation}
\frac{1}{Z(\phi,m)}\phi^{K_d(\succ^*,\succ)}, 
\label{eq:mallows-generator}
\end{equation}
where $Z(\phi, m)$ is a normalization constant. 

A random permutation of $A$ is generated to serve as to ground truth ranking $\succ^*$. We then use Equation~\ref{eq:mallows-generator} to generate {\bf task rankings} over models, $\succ^*_v$, where each task ranks models. Intuitively, if $\phi$ is close to 0, then we are more likely to draw task rankings which are close to $\succ^*$ as measured by Kendall-tau distance. As $\phi$ approaches 1, we are more likely to draw a more diverse set of task rankings since the effect of the Kendall-tau distance on the probability of any ranking being generated is smaller.

For each task ranking $\succ^*_v$, $m$ distributions of scores, $S_{v,a}$ for $a \in A$ mentioned in Section~\ref{sec:active-eval}, are generated by sampling uniform random numbers in a bounded score internal and assigning the values as the means of $S_{v,a}$ ($\mu_{v,a}$), such that they are sorted according to $\succ^*_v$. Each $S_{v,a}$ is then normally-distributed according to $\cN(\mu_{s,a}, \sigma)$.
A summary of the process is depicted in Figure~\ref{fig:synthetic-data-gen}.

\subsubsection{Plackett-Luce}
\label{sec:plackett-luce}

We also try an alternative synthetic data model: 
Plackett-Luce. Results are consistent with those on Mallows; the description is similar, and is presented in Appendix~\ref{app:plackett-luce}.

\section{Algorithms for Active Evaluation}
\label{sec:algorithms}

We now describe several algorithms which we will compare experimentally in the following section (summarized in Table~\ref{tab:algorithm-summary}).
Algorithms for active evaluation are naturally decompose into three steps (with line references to Algorithm~\ref{alg:active-eval-loop}):
\begin{enumerate}
    \item the \defword{task selection} chooses a task $v^t \in V$ (line~\ref{alg:selection}),
    \item the \defword{agent selection} chooses two agents  $(a^t_i, a^t_j)$ (line~\ref{alg:selection}), and
    \item the process that updates tracked information based on the score sample $s^t$ (line~\ref{alg:update}).
\end{enumerate}

\subsection{Baselines and Bandit Approaches}
\label{sec:basic-baselines}

\begin{table*}[t!]
\begin{tabular}{c|ccl}
{\bf Name} & {\bf Task Selection} & {\bf Agent Selection} & {\bf Update}\\
\hline
\textsc{UniformAveraging} & Uniform Random & Uniform Random & Track cumulative average $\bar{S}_a$ per agent (aggregated across tasks)\\
% \textsc{TaskAverageBorda} & Uniform Random & Uniform Random & Track per-task score average $\bar{S}_{v,a}$; Borda-aggregated task rankings \\
\textsc{BasicUCB} & Uniform Random & Agent-level UCB & Update means and visit counts for agent-level bandit\\
\textsc{KemenyEl} & Uniform Random & Uniform Random & Update $Q$ matrix and statistics for pair of agents selected\\
\hline
\textsc{BatchElo} & Uniform Random & Uniform Random & Track cumulative wins, losses and draws; rank by Elo ratings\\
\textsc{BatchCopeland} & Uniform Random & Uniform Random & Growing-batch Copeland voting~\citep{Copeland50} \\
\textsc{BatchRankedPairs} & Uniform Random & Uniform Random & Growing-batch Ranked Pairs voting~\citep{Tideman87} \\
\textsc{BatchMaximalLotteries} & Uniform Random & Agent Strategy & Growing-batch Iterative Maximal Lotteries voting~\citep{lanctot2023evaluating} \\
\textsc{BatchSCO} & Uniform Random & Uniform Random & Growing-batch version of SCO~\citep{Lanctot25SCO} \\
\textsc{BatchNashAveraging} & Task Strategy & Agent Strategy & Growing-batch version of Nash Averaging~\citep{Balduzzi18ReEval} \\
\hline
\textsc{MeanModelCopeland} & Uniform Random & Uniform Random & Estimate task rankings via $\bar{s}_{v,a}$ and return VasE(Copeland)\\
\textsc{MeanModelRankedPairs} & Uniform Random & Uniform Random & Estimate task rankings via $\bar{s}_{v,a}$ and return VasE(Ranked Pairs)\\
\textsc{MeanModelMaxLot} & Uniform Random & Uniform Random & Estimate task rankings via $\bar{s}_{v,a}$ and return VasE(Iterative Max Lot)\\
\textsc{PropRepresentation} & Greedy Prop. Rep. & Uniform Random & Online version of Greedy Metritocracy algorithm~\citep{procaccia2025metritocracy} \\
\hline
\textsc{OnlineElo} & Uniform Random & Uniform Random & Track ratings only; update ratings from each game outcome\\
\textsc{OnlineSCO} & Uniform Random & Uniform Random & Track and update ratings only (analogously to online Elo)~\citep{Lanctot25SCO} \\
\textsc{OnlineMaximalLotteries} & Uniform Random & Agent Strategy & Fully-online Maximal Lotteries (based on adersarial bandits)~\citep{Fishburn84} \\
% \textsc{KemenyEl} & Uniform Random & Uniform Random & Fully-online Kemeny ranking elicitation~\citep{George24KemenyEl} \\
\textsc{OnlineNashAveraging} & Task Strategy & Agent Strategy & Fully-online Nash Averaging (based on adversarial bandits)~\citep{Balduzzi18ReEval} \\
\hline
\end{tabular}
\caption{A summary of algorithms for active evaluation. For methods based on zero-sum games, the Task Strategy and Agent Strategy are sampling distributions in the underlying game, \ie with uniform exploration mixed in: \eg $\sigma'(a) = \frac{\epsilon}{|A|} + (1-\epsilon)\sigma$.}
\label{tab:algorithm-summary}
\end{table*}

The most basic procedure for steps (1) and (2)  is \textsc{UniformRandom} which samples $v \sim \textsc{Uniform}(V)$ uniformly at random and $(a_i, a_j) \sim \textsc{Uniform}(A)$ without replacement.
The simplest update rule tracks the cumulative average score $\bar{S}_a$ aggregated across all tasks and outputs a ranking $\succ^t$ by sorting $\bar{S}_a$ for $a \in A$. Denote this simplest baseline algorithm \textsc{UniformAveraging}.

A second baseline, \textsc{BatchElo} is based on the Elo rating system: it also samples tasks and agents uniformly randomly, and maintains a historical record of win/loss outcomes where $s(v^t, a^t_i) > s(v^t, a^t_j)$ counts as a win for $a_i$ and loss for $a_j$. Then an efficient batch implementation of Elo~\cite{Hunter04btmodels} computes the best fit for ratings $\btheta$ and $\succ^t$ is obtained by sorting the ratings in descending order.
We also use \textsc{OnlineElo}, the standard update typically used on competitive gaming platforms from single-game outcomes.

\subsubsection{Stochastic Bandit-based Approaches}
\label{sec:ucb}

We suggest  bandit-based approaches based on Upper Confidence Bounds (UCB) algorithm~\citep{Auer02UCB} as often used in Monte Carlo tree search~\cite{Kocsis06UCT}.

\textsc{BasicUCB}  is a single multi-armed bandit with one arm assigned to each agent (scores aggregated across tasks. 
Each agent $a \in A$ is assigned a UCB score:
\[
\textsc{UCBScore}(a, T) = \bar{s}_a + C \sqrt{ \log(2T) / T_a},
\]
where $\bar{s}_a$ is the average score of agent $a$ , $C$ is a constant set to $\sqrt{2} \Delta_S$, $2T$ is the total number of updates and, $T_a$ is the number of updates to agent $a$.
Tasks are selected uniformly at random. The final ranking is a sorting of visit counts per arm.

\subsubsection{Dueling Bandit Approaches}

When using a dueling bandit~\cite{Yue12CopelandDB} approach feedback comes in the form of a preference between the two arms pulled. Dueling bandits have been used for ranking, for example Copeland dueling bandits~\cite{zoghi2015copeland}.
We use a recent dueling bandit based on Kemeny ranking called KemenyEl~\cite[Algorithm 1]{George24KemenyEl}. Like UCB, KemenyEl uses statistics to estimate the noise between pairwise preferences, factoring it into its decision rule. Unlike UCB, the Kemeny rule is used for ranking.

We adapt KemenyEl to the online evaluation setting in the following way.
The approximation accuracy of KemenyEl holds with probability $1-\delta$, where $\delta$ is chosen by the user and determines the number of samples per pairwise matchup.
In the online setting, we first choose an initial $\delta_0 = 0.1$; once the number of samples per matchup is reached (\ie one matchup per round of Alg. ~\ref{alg:active-eval-loop}), a new epoch is started ($k \leftarrow k+1$) and we set $\delta_k = \frac{\delta_{k-1}}{2}$. 
Similarly, we set the acceptable Kendall-tau distance error, $\rho_0$, to be half of the total Kendall-tau distance and then halve it every epoch.

\subsection{Growing Batch and Online Algorithms}

Any algorithm that is normally run offline can be run online by simply collecting evaluation data over time and running the algorithm on the data set on iteration $t$. 
We call this the {\bf growing-batch} version of an offline algorithm.
VasE methods interpret evaluations as data sets similarly to how Elo does: for example, given two score samples for task $v$ if, $s(v^t, a^t_i) > s(v^t, a^t_j)$, this counts as a vote $a_i \succ a_j$.
Given the pairwise preference collection, we examine several Condorcet methods: \textsc{BatchCopeland}, \textsc{BatchRankedPairs}, and \textsc{BatchMaximalLotteries} corresponding to growing-batch versions of the Copeland, Ranked Pairs, and Maximal Lotteries voting schemes~\citep{Copeland50,Tideman87,Fishburn84,lanctot2023evaluating}.
We use an iterative variant of Maximal Lotteries~\citep{lanctot2023evaluating} which provides a ranking over agents.

We also use a batch version of Soft Condorcet Optimization~\citep{Lanctot25SCO}, which is a gradient descent algorithm run on a smooth Kendall-tau loss and its fully-online version which is analogous to Elo's standard online update.
We introduce a new fully-online version of Maximal Lotteries that we describe separately in Section~\ref{sec:online-ml-nashavg}.

A different way to apply VasE algorithms online is to use a {\bf mean model}. Mean-model VasE algorithms maintain $nm$ emprical averages of each score $\bar{s}_{v,a}$. Estimated per-task agent rankings $\hat{\succ}_v$ are obtained by sorting $\bar{s}_{v,a}$ for $a \in A$; then, a voting rule returns a ranking from the $n$ ranked-ballot votes.
\textsc{ProportionalRepresentation} is a mean-model adaptation of the recently-proposed Greedy Metritocracy algorithm~\citep[Algorithm 1]{procaccia2025metritocracy}. In particular, it uses estimated task rankings $\hat{\succ}_v$ as votes to determine an {\it reduced task set} based on proportional representation, recomputing this subset on each iteration based on updated estimates.
The group size parameter ($g = \lceil g_f \cdot n \rceil$) captures the trade-off between the granularity of representation and the size of subset needed to satisfy positional representation.
Tasks are selected uniformly at random from this reduced task set, with a small probability (0.05) of choosing uniformly form the entire set of tasks to ensure exploration. The reported ranking is then obtained by running Ranked Pairs votes over the votes from estimated task rankings.

% \mlnote{Describe burn-in}
All of the mean-model and growing-batch algorithm use a {\bf burn-in} phase: for the first $nm$ iterations, the task and agent sampling one cycle through a uniform shuffling of the unique $(v, a) \in V \times A$ pairs.
This burn-in phase has been shown to benefit stochastic gradient descent algorithms~\cite{roux2013stochastic}.
Data is still collected and rankings reported in the same way during the burn-in period. 

\subsubsection{Growing Batch Nash Averaging}
\label{sec:game-theory-eval}

One way to define ratings of players is to analyze the one-shot meta-game played between agents and tasks~\cite{Balduzzi18ReEval}.
This matrix game is a two-player zero-sum game where the row player chooses an agent $a \in A$ and the column player chooses a task $v \in V$. 
The utility to the row player for the joint choice $(a, v)$ is the score $s_{v,a}$ and the utility to the column player is $-s_{v,a}$. In the original paper, these utilities where averaged over many samples, so utilities took the form
\eg $(\bar{s}_{v,a}, -\bar{s}_{v,a})$. 
The rating of agent $a$, $\theta_a$, is defined as the expected utility of strategy $a$ versus the task player's Nash equilibrium strategy, which can be computed efficiently (polynomial time) since the game is two-player zero-sum.
A growing-batch version employs Nash averaging on the batch of data collected so far: utilities are averages, and it computes ratings as in the original Nash averaging. A ranking is then obtained by sorting the ratings.

Nash averaging can be also run fully online by using adversarial bandits. We elaborate on this in the next section.

\subsubsection{Online Max. Lotteries and Nash Averaging}
\label{sec:online-ml-nashavg}

Similarly to Nash averaging, the Maximal Lotteries voting method also involves solving a two-player zero-sum ``margin game'' and using its solution to define the set of winners.
The game is an two-player zero-sum agent-vs-agent game: the utility to the row player for strategy $a_i$ versus the column player's choice $a_j$ is the voter margin $\delta(a_i, a_j) = N(a_i, a_j) - N(a_j, a_i)$, where $N(a_i, a_j)$ is the number of votes where $a_i \succ a_j$.
As above, the growing-batch version of Maximal Lotteries derives this game from the growing data set; like Nash averaging, it can also be run fully online.

We defer the details of solving two-player zero-sum games using adversarial bandits to Appendix~\ref{app:games-adv-bandits}.
We base our implementations on regret-matching~\cite{Hart01}.
In Nash averaging, when agent $i$ is sampled against task $v^t$ the utility is $s(v^t, a_i)$ (obtained in Line~\ref{alg:sample-scores} of Algorithm~\ref{alg:active-eval-loop}).
The solution to the game (approximate Nash equilibrium) is extracted as a pair of distributions (stochastic strategies) $\sigma = (\sigma_1, \sigma_2)$ for players 1 and 2. 
In the case of Nash averaging, $\sigma_1 \in \Delta(A)$ and $\sigma_2 \in \Delta(V)$.
In the case of Maximal Lotteries, $\sigma_1, \sigma_2 \in \Delta(A)$ where $\Delta(A)$ represents a simplex over elements in $A$.
The utility for the row player choosing $a_i$ playing against column player choosing $a_j$ when task $v^t$ was sampled (uniformly) is:
$\hat{u}_i(v^t, a_i, a_j) = 2 \cdot \mathbb{I}[s(v^t, a_i) > s(v^t, a_j)] - 1$,
where $\mathbb{I}$ is the indicator function.
In both cases (Nash averaging and Maximal Lotteries), to extract a ranking, we define the ratings as expected values against the equilibrium strategy; \eg for $a \in A$: $\textsc{Rating}(a) = \E_{b \sim \sigma_2} u_1(a, b)$,
where $u_1(a, b)$ is the average utility to player 1 given choices $a$ and $b$ as described in the games above. The rankings are then extracted by sorting the ratings. This naturally extends Maximal Lotteries' distribution over winner agents to a ranking among agents~\citep{marris2022gametheoreticratingnplayer}.

\subsection{Summary of Known Convergence Results}

Since the game online Nash Averaging plays is a sample-based version of the true underlying game, the average strategy used by adversarial bandits is guaranteed to converge to an approximate equilibrium with high probability, and the error decreases as $\epsilon_t = O(1/\sqrt{t})$ where $t$ is the number of iterations~\cite{Auer98Exp3}, and ratings will differ from the true ratings by at most $\epsilon_t$. Similarly, the solution computed by Batch Maximal Lotteries is obtained by solving a two-player zero-sum margin game derived directly from a preference profile. The error of the estimated expected utility of a cell in the margin game (defined in Section~\ref{sec:online-ml-nashavg}) decreases as $O(1/\sqrt{t})$ and hence the Maximal Lottery solution we derive from it will be a 
$O(1/\sqrt{t})$-approximate solution. 

Asymptotic guarantees on GRE requires knowing the ground truth and, possibly, also the shape of the data both of which are generally not known. When data is shaped as described in Sec~\ref{sec:mallows}, the mean model variants’ estimated means converge to the true means of the underlying score distributions at a rate of $1/\sqrt{n}$ by the Monte Carlo square root law; consequently, the estimated task rankings will converge to the true task rankings and the overall ranking to a VasE-aggregated one from the task rankings. 

\section{Experimental Results}
\label{sec:experiments}

\begin{figure}[ht!]
\begin{tabular}{c}
\includegraphics[width=0.392\textwidth]{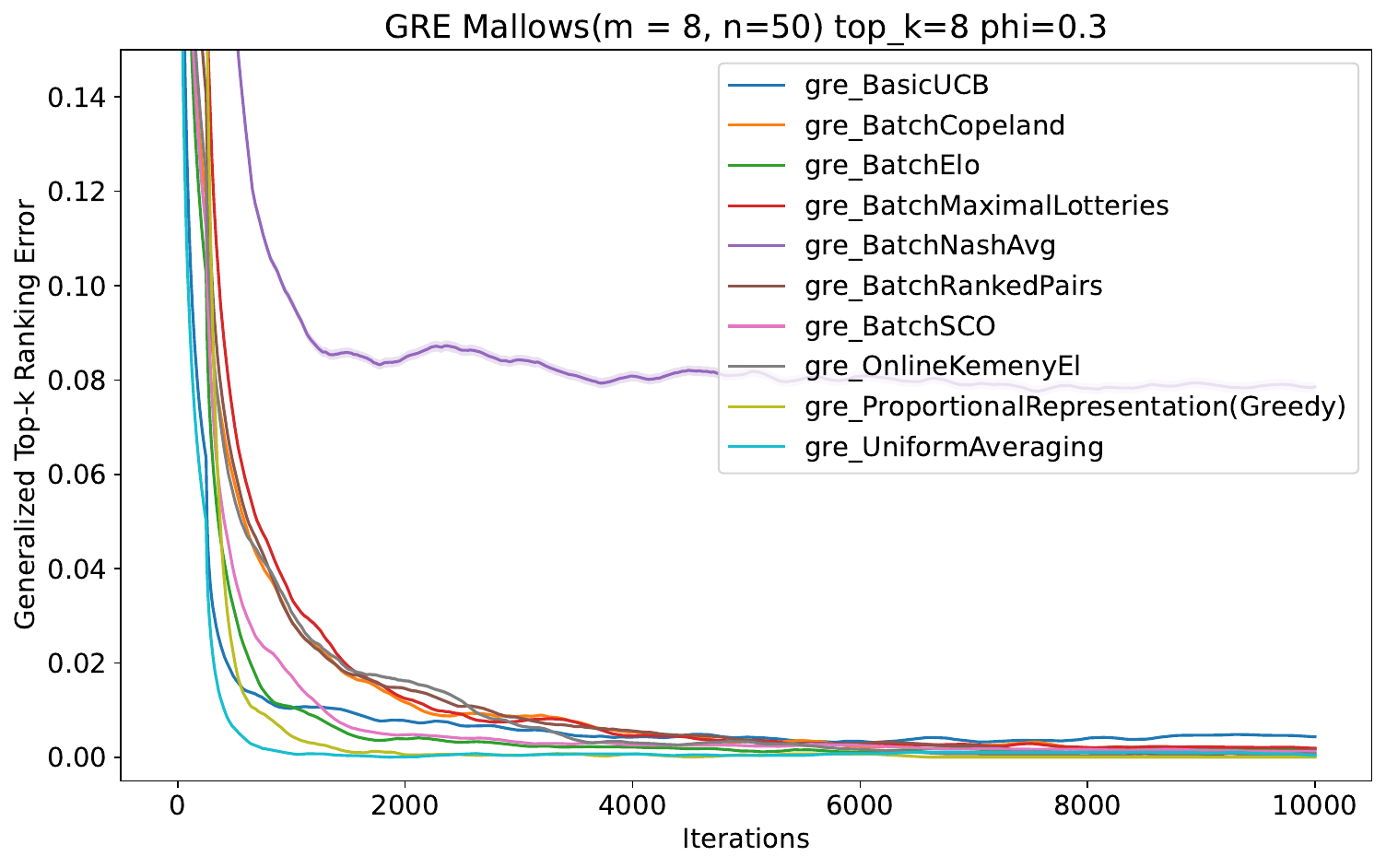} \\
\includegraphics[width=0.392\textwidth]{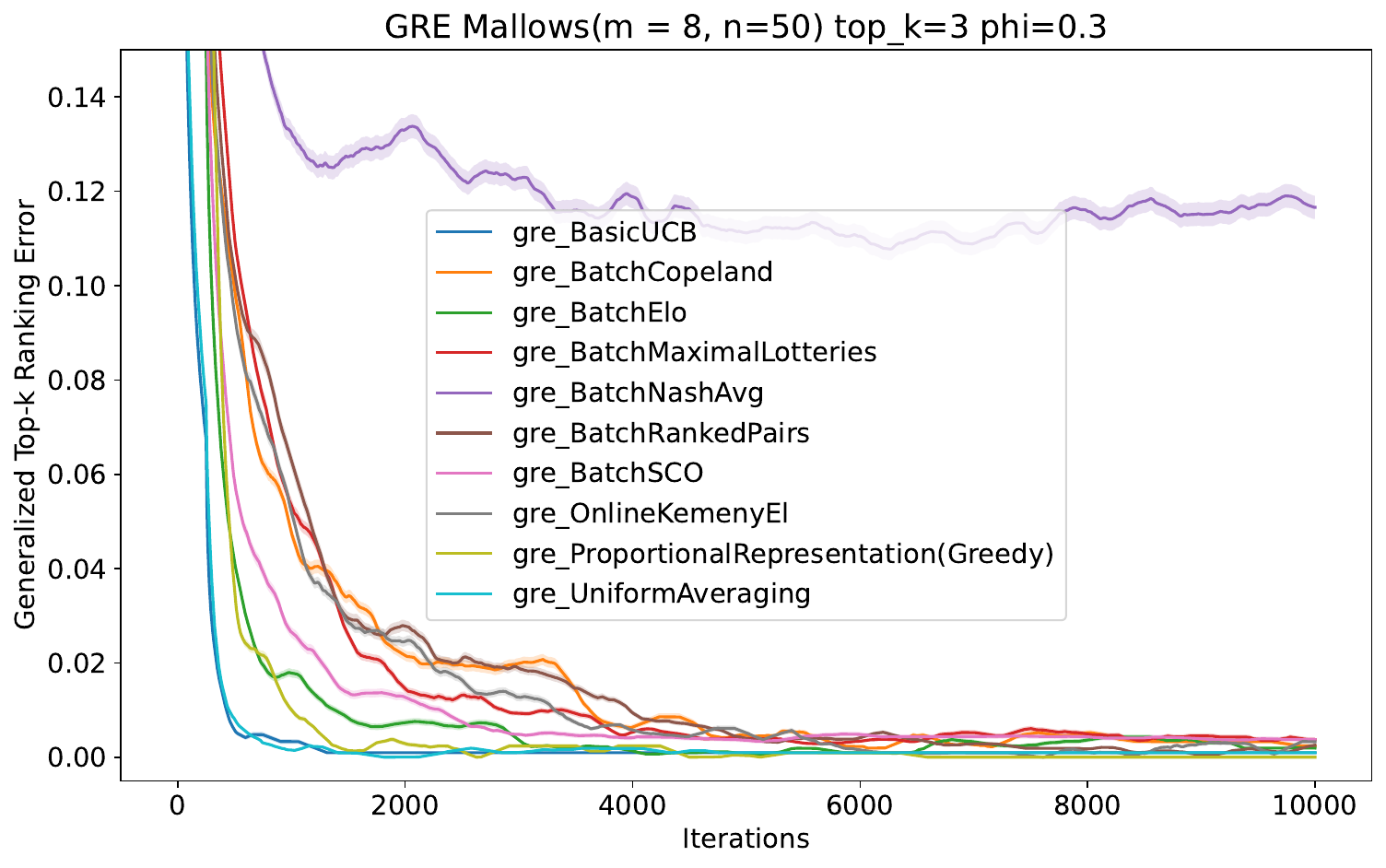} \\
\includegraphics[width=0.392\textwidth]{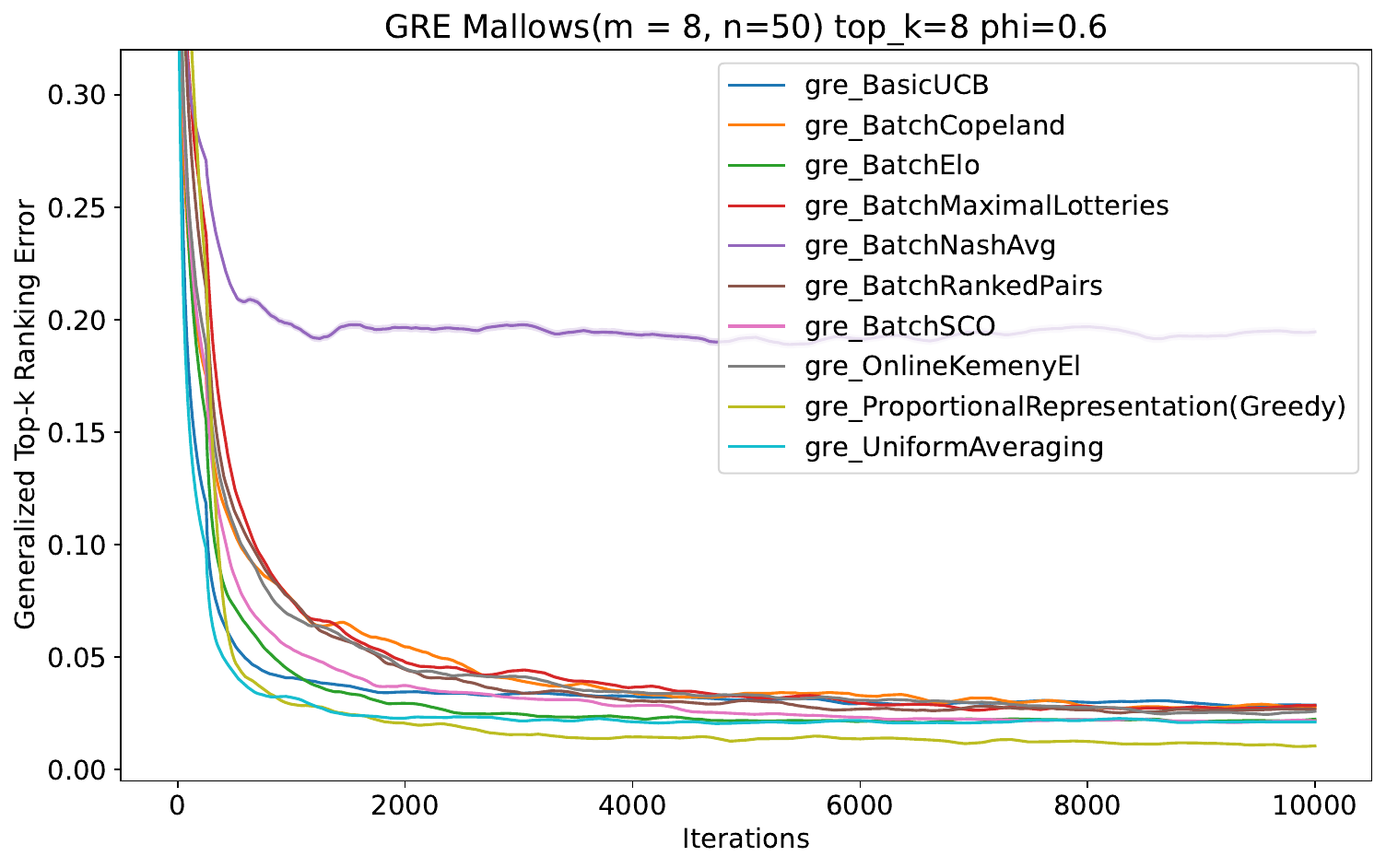} \\
\includegraphics[width=0.392\textwidth]{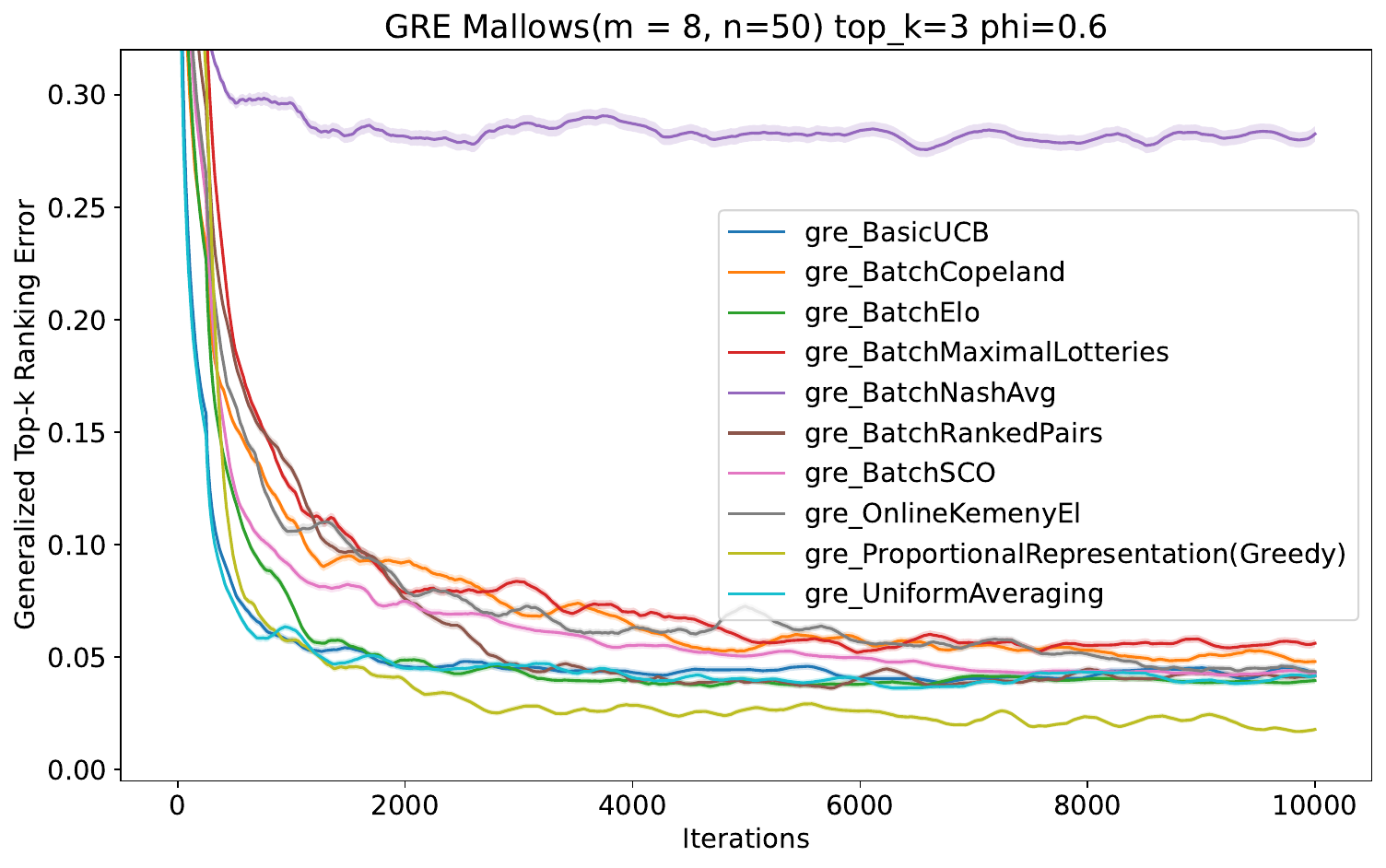} \\
\end{tabular}
\caption{Exp Condition 1: Generalized top-$k$ ranking error (GRE) over iterations for several algorithms. Each line is a mean over the same 100 seeds with 95\% confidence intervals in shaded area. Each data point is a sliding-window average of the most recent 250 values of GRE for the reported ranking $\succ^t$ at iteration $t$. The top two graphs use dispersion $\phi=0.3$, with $k \in \{3, 8\}$, where $k=8$ corresponds to normalized Kendall-tau distance. The bottom two use $\phi=0.6$ and vary $k$ similarly.}
\label{fig:expcond1-rank-error}
\end{figure}

We now compare the performance of each active evaluation algorithm, first on synthetic data then on real Atari agent data.

\subsection{Experiments with Synthetic Data}

For experiments using synthetic data using models described in Section~\ref{sec:synthetic-data}.
For score distributions, means are drawn from the interval $\mu_{v,a} \in [0, 100]$ and use standard deviation of $\sigma = 20$.

\subsubsection{Experimental Condition 1: Ranking Error Reduction Rate}
\label{sec:expcond1}

A primary motivation of research on active evaluation is designing algorithms that use/sample data efficiently, so a main research question is how well algorithms can report a ranking that is close to the ground truth as a function of number of samples. 
We use a moderately sized problem with $m=8$ agents and $n=50$ tasks; each seed determines a new ground truth ranking, task rankings, and score distributions.
Since the fully-online variants of algorithms are approximating the growing-batch versions, we restrict this first experiment to the growing batch algorithms with basic baselines and bandit approaches.
Figure~\ref{fig:expcond1-rank-error} shows the generalized top-$k$ ranking error (GRE, definition~\ref{def:gre}) as a function if iterations of Algorithm ~\ref{alg:active-eval-loop}.

Our first observation is that, when $k = 3$, it is possible to drive the error to 0 within 2000 iterations and several algorithms achieve this. Our second observation, \textsc{UniformAveraging}-- the most basic of baselines-- performs particularly well in all settings, and in three of four cases is reduces the error fastest in the first 1000 iterations, though \textsc{BasicUCB} has comparable performance when $\phi=0.6$. In general, \textsc{UniformAveraging}, \textsc{BasicUCB}, \textsc{BatchElo}, \textsc{BatchSCO}, and \textsc{ProportionalRepresentation} reduce the error most quickly at the beginning.
In Nash averaging, we observe a phenomenon similar to ``adversarial task selection'' reported in~\citep{lanctot2023evaluating}: since it is a zero-sum game of agents versus tasks, the task player prefers to choose tasks where the agents' utilities are minimized, resulting in an equilibrium distribution skewed towards task rankings that may be arbitrarily far from the ground truth. Hence, the ranking overfits to these adversarially-chosen tasks.
When $\phi=0.6$, \textsc{ProportionalRepresentation} is a clear winner, deviating significantly from the other methods. We try group size proportions $g_f \in \{ 0.1, 0.2, 0.25, 0.3, 0.35, 0.5, 0.6, 0.75 \}$, finding that $g_f = 0.1$ to be the best value and use it by default. 
We show trade-offs between group sizes, reductions in task sets, and GRE in Appendix~\ref{sec:proprep-group-sizes}. 

In Table~\ref{tab:expcond1-tab-agre}, we report on the longer-term performance in the form of average GRE at $t=10000$ for $\phi = 0.3$ and $k \in \{ 3, 8 \}$, which can be interpreted as an approximate average area-under-the-GRE-curve. We observe that \textsc{BasicUCB} and \textsc{UniformAveraging} peform best when $k=3$ and \textsc{UniformAveraging} performs best when $k = 8$. In both cases, \textsc{BatchElo} also reduces the rank error at a fast rate.
\begin{table}
\begin{tabular}{c|ccc}
\multirow{2}{*}{\bf Algorithm}     & \multicolumn{3}{c}{ $\textsc{AGRE}(k, \phi)$ } \\
                                  & $(3, 0.3)$ & $(8, 0.3)$ & $(3, 0.6)$\\
\hline
\textsc{BasicUCB}                   & {\bf 0.003}   & 0.007        & 0.046 \\
\textsc{UniformAveraging}           & 0.003         & {\bf 0.002}  & 0.049 \\
\textsc{ProportionalRepresentation} & 0.009         & 0.005        & {\bf 0.037} \\
\textsc{BatchElo}                   & 0.009         & 0.006        & 0.050 \\
\textsc{BatchSCO}                   & 0.012         & 0.007        & 0.063 \\
\textsc{KemenyEl}                   & 0.017         & 0.010        & 0.072 \\
\textsc{BatchCopeland}              & 0.018         & 0.010        & 0.072 \\
\textsc{BatchMaxLot}                & 0.019         & 0.013        & 0.079 \\
\textsc{BatchRankedPairs}           & 0.020         & 0.012        & 0.063 \\
\textsc{BatchNashAvg}               & 0.124         & 0.088        & 0.285 \\
\hline
\end{tabular}
\caption{Exp. Condition 1: Average Generalized Top-$k$ ranking error, $\textsc{AGRE}(\succ^{1:t}, \succ^*, t, k = \cdot)$, at $t = 10^4$ and various values of $k$ and $\phi$. All values have 95\% confidence intervals $< 10^{-3}$.}
\label{tab:expcond1-tab-agre}
\end{table}

\subsubsection{Experimental Condition 2: Batch versus Online}
\label{sec:expcond2}

Firstly, we see that there are several methods that reach 0 error in less than 10000 iterations: \textsc{BatchSCO}, \textsc{BatchElo}, \textsc{OnlineElo}, \textsc{ProportionalRepresentation}, and all of the mean-model variants.
Second, while the batch version of Maximal Lotteries reduces rank error faster than its online variant, in Nash averaging this is reversed, possibly due to the average strategy in the adversarial bandit changing more smoothly.
Third, mean-model variants and \textsc{ProportionalRepresentation} reduce error rate faster than the batch versions of these algorithms, several reaching 0 error in less than 6000 iterations (\textsc{MeanModelMaximalLotteries} and \textsc{MeanModelRankedPairs}).

\begin{figure}[t!]
\begin{tabular}{c}
\includegraphics[width=0.39\textwidth]{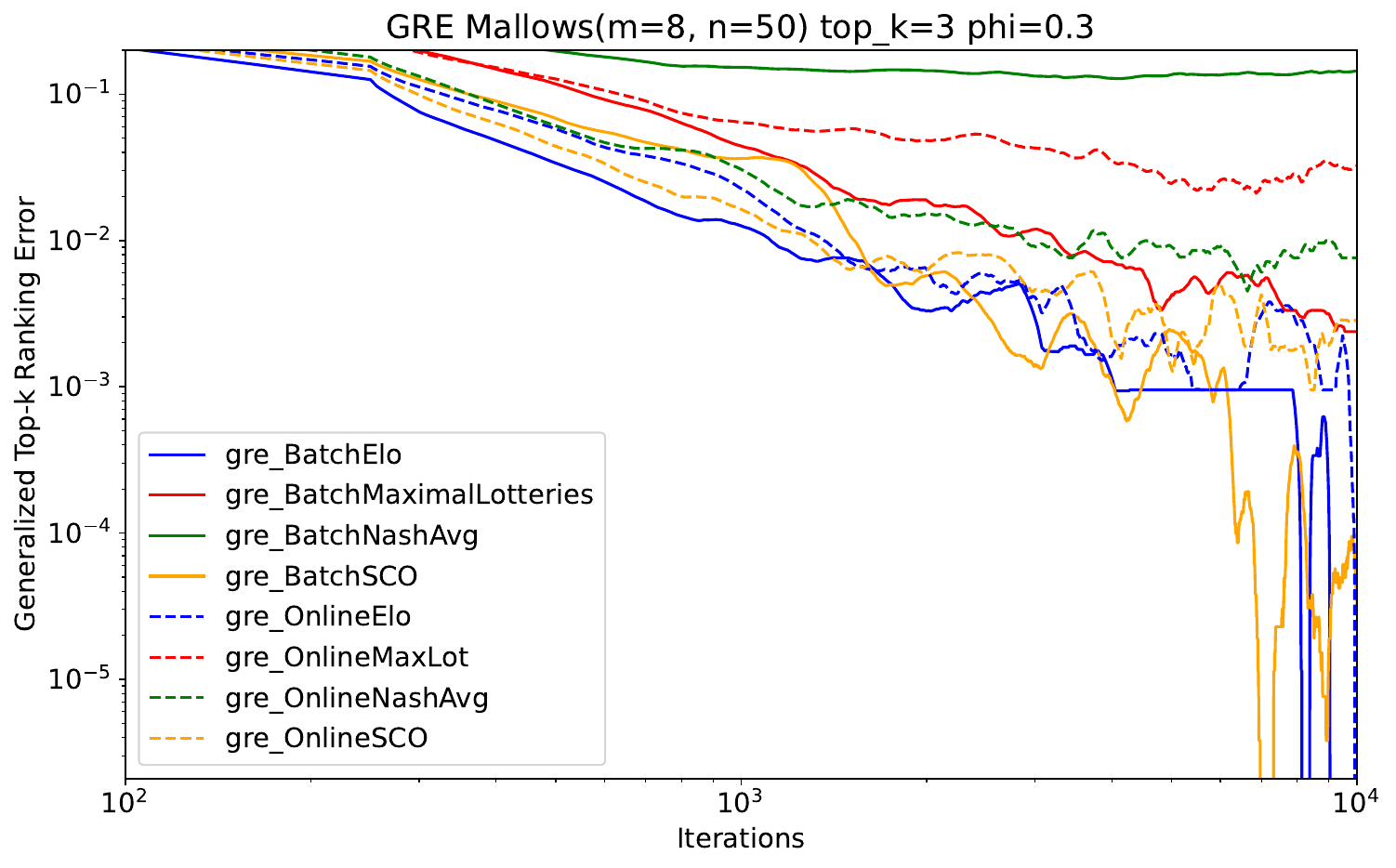} \\
\includegraphics[width=0.39\textwidth]{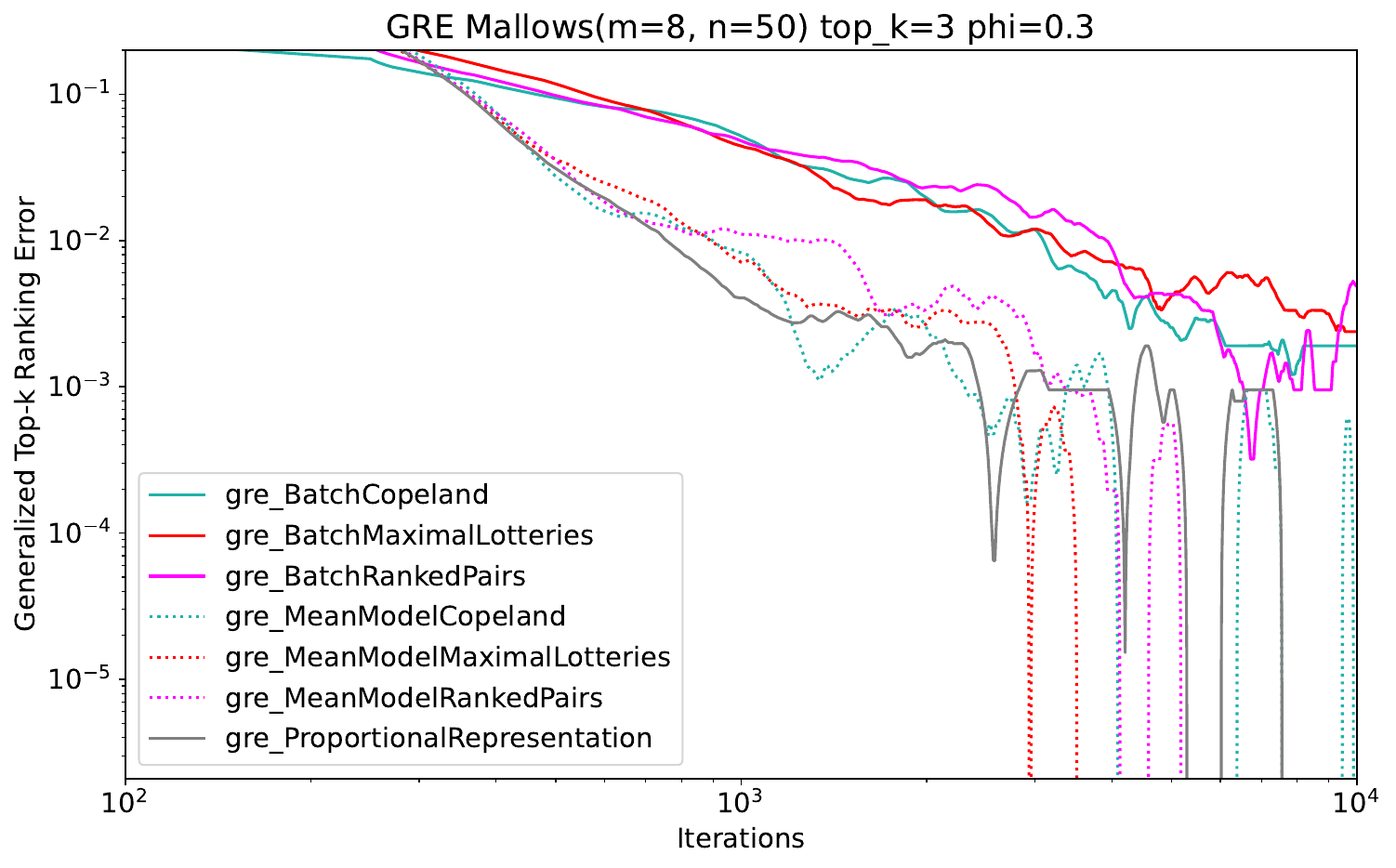}
\end{tabular}
\caption{Exp Condition 2: Generalized top-$k$ ranking error (GRE) over iterations for several algorithms. Each data point is a sliding-window average of the most recent 250 values of GRE for the reported ranking $\succ^t$ at iteration $t$. Both graphs use $k=3$. Please note the log scale for both axes.}
\label{fig:expcond2-batch-vs-online}
\end{figure}

\subsection{Experiments with Real Data}

% Main idea here: does theory match practice? If not, how does it differ?
In this section, we simulate the online sampling of evaluation data by presenting theses data sets as data generators. %similarly to how the synthetic data is obtained in Secion~\ref{sec:synthetic-data}.

\subsubsection{Atari}
\label{sec:atari}

We use previos results of agents playing all 57 classic Atari games via the Arcade Learning Environment (ALE)~\citep{bellemare13arcade}. 
In particular, we use the data from the ``Agent57 paper'' table in~\citep[Section H.4]{badia2020agent57outperformingatarihuman} which comprises $m=8$ agents and $n=57$ tasks (Atari games).
The agents are varied among uniform random, human expert, and several deep reinforcement learning algorithms.
In particular, unlike much of the previous work on deep reinforcement learning in ALE, this table reports both mean scores $\mu_{v,a}$ {\it and standard deviations} $\sigma_{v,a}$. Hence, we simulate a data generator by setting the score distributions $S_{v,a} = \cN(\mu_{v,a}, \sigma_{v,a})$ where $\cN$ is a normal distribution, and otherwise identically to the synthetic data generators. 
Since the scale of rewards is different in each game, we also linearly normalize the scores per game from range $[\min_{a \in A} \mu_{v,a}, \max_{a \in A} \mu_{v,a}]$ to within range $[0, 100]$. This score normalization does not affect the task rankings, is required by Nash averaging~\citep{Balduzzi18ReEval} and averaging methods, and helps maintain a consistent interval for exploration constant in bandit methods. 
In the absence of a proper ground truth we compute the ranking that minimizes the average Kendall-tau distance to the data: $\succ^* = \min_{\succ'} \frac{1}{57} \sum_{v \in V} K_d(\succ', \succ_v)$; this ranking is a maximum likelihood estimator of the ground truth under a Mallows model~\cite[Chapter  8]{Brandt16Handbook}. We also show in Appendix~\ref{app:kemeny-ground-truth} that it also recovers the ground truth ranking on synthetic data.

The average generalized top-$k$ ranking error is summarized in Table~\ref{tab:atari-agre} (for full graphs of GRE over iterations, see Figure~\ref{fig:atari-gre-graphs} in Appendix~\ref{app:additional-results}). 
In this setting, \textsc{BatchSCO} performed best followed by \textsc{OnlineSCO} and five of the top six were VasE methods.
The order of the simple baselines \textsc{BasicUCB} and \textsc{UniformAveraging} is completely reversed from the synthetic data setting: they perform worst in Atari; this indicates that in some domains one may need to be careful when aggregating scores across tasks. 
Nash averaging also performs poorly, which could also be due to the adversarially-chosen task distribution. We do note that there are $\frac{15}{57}$ tasks whose rankings different significantly from the ground truth, \ie where $K_d(\succ^*_v, \succ^*) \ge 8$; skewing a task distribution toward any of these high-distance tasks could significantly hinder the ability to find the ground truth ranking. 
We compare the task variation from the ground truth to samples from the synthetic data in Appendix~\ref{app:data-variation}.

% top_k = 3
% agre_ProportionalRepresentation	0.103471    0.000553

% top_k = 8
% agre_ProportionalRepresentation	0.033564	0.000151

\begin{table}
\begin{tabular}{c|cc}
{\bf Algoirthm} & $\textsc{AGRE}(k=3)$ & $\textsc{AGRE}(k=8)$ \\
\hline
\textsc{BatchSCO}              & {\bf 0.019031} &  {\bf 0.008095} \\
\textsc{OnlineSCO}             & 0.025931	    &  0.009034 \\
\textsc{KemenyEl}              & 0.027322       &  0.011631  \\
\textsc{BatchRankedPairs}      & 0.027873       &  0.012952  \\
\textsc{BatchElo}              & 0.032967       &  0.010907  \\
\textsc{BatchCopeland}         & 0.037158       &  0.013470 \\
\textsc{OnlineElo}             & 0.037259       &  0.012504  \\
\textsc{MeanModelCopeland}     & 0.042865       &  0.013617 \\
\textsc{BatchMaximalLotteries} & 0.043418       &  0.021718  \\
\textsc{MeanModelMaxLot}       & 0.045105       &  0.013727 \\
\textsc{ProportionalRepresentation} & 0.045472  &  0.013558 \\
\textsc{MeanModelRankedPairs}  & 0.048279       &  0.014740  \\
\textsc{OnlineMaxLot}          & 0.135168       &  0.057263  \\
\textsc{OnlineNashAvg}         & 0.240771       &  0.105709  \\
\textsc{BasicUCB}              & 0.241976       &  0.074383 \\
\textsc{UniformAveraging}      & 0.242354       &  0.259529 \\
\hline
\end{tabular}
\caption{Atari: Average Generalized Top-$k$ ranking error, $\textsc{AGRE}(\succ^{1:t}, \succ^*, t, k = \cdot)$ at $t = 10^4$ on simulated Atari data generator. 95\% confidence intervals for each entry was $< 10^{-3}$.}
\label{tab:atari-agre}
\end{table}

\section{Related Work}
\label{sec:relwork}

Most of the related previous work discussed in this paper (such as VasE and game-theoretic evaluation) are based on the standard offline case, \ie run on a batch of data. However, given the immense costs of evaluating language models, there are some recent works studying active evaluation for language models.

Li et al.~\citep{li2024active} explicitly models dependencies in the test examples allowing predictions for remaining examples, and uses reinforcement learning to learn a policy to select subsets.
Their method adaptively selects different subsets of prompts per model based on each model's strengths and weaknesses.
Angelopoulos et al.~\citep{angelopoulos2025costoptimal} propose a cost-optimal framework for how to balance obtaining evaluation data from strong-but-costly raters versus weak-but-cheap raters based on prediction-powered inference. Costs per evaluation are explicitly assumed and a main result is an optimal policy to actively sample prompts.
In contrast, our conceptual framing is not specifically intended for (however, compatible with) language model evaluation and focuses on task and model selection rather than LM prompt subselection and trade-offs in rater quality. 
%However, (multi-task) language model evaluation does fit into our proposed framework, where score distributions come from prompt collections per task.

There is work in computational social choice that examines the sample complexity required for noisy votes to reveal the truth, under what conditions, and algorithms which are guaranteed to find the ground truth in the limit of infinite samples~\citep{Caragiannis16When}.

\section{Conclusion and Future Work}

In this paper, we formally described the problem of active evaluation for general (multi-task) agents and proposed data generation techniques.
% We defined generalized metrics for capturing the correctness of reporting rankings compared to ground truth ranking. We compared the performance of fourteen algorithms ranging from simple baselines, classical methods, and recent methods based on social choice theory and game theory. 
We found that across synthetic data and real data experiments, \textsc{BatchElo} is a consistently reliable choice for obtaining low ranking error across samples, performing especially well in the low task variation ($\phi = 0.3$) regime.
However, \textsc{BatchSCO} and \textsc{OnlineSCO} are comparable and come in as a close second in synthetic data, while outperforming \textsc{BatchElo} two-fold in Atari agent evaluation.
In the synthetic data experiments, \textsc{BatchElo} was outperformed by \textsc{UniformAveraging} and \textsc{BasicUCB} but they both performed poorly on the Atari experiment, and Nash averaging also performing poorly; evaluators may need to exercise caution when aggregating or comparing score values across tasks.
\textsc{BatchElo} was significantly outperformed by \textsc{BatchSCO}, \textsc{OnlineSCO}, and \textsc{BatchRankedPairs}, with SCO being the clearly best method on the Atari experiment.
In Atari (unlike in the synthetic data), batch variants seemed to outperform their fully-online versions; however, online and mean-model variants are still viable choices that offer lower computational complexity. 

Future work could derive formal properties and statistical approximation guarantees as a function of the number of rounds.
Also, only a few of the algorithms we compared adaptively modified the task selection. We hope that future work investigates the benefit of adaptive sampling strategies for data collection that use confidence intervals, uncertainty sampling, expected model change, error reduction, or task/agent score correlations~\cite{Settles09,yang2018benchmark,George24KemenyEl}.

%%%%%%%%%%%%%%%%%%%%%%%%%%%%%%%%%%%%%%%%%%%%%%%%%%%%%%%%%%%%%%%%%%%%%%%%

\begin{acks}
We would like to thank Manon Revel for the helpful feedback on a previous draft of the paper and Serena Wang for help with the Metritocracy (proportional representation) algorithm.
\end{acks}

%%%%%%%%%%%%%%%%%%%%%%%%%%%%%%%%%%%%%%%%%%%%%%%%%%%%%%%%%%%%%%%%%%%%%%%%

%%% The next two lines define, first, the bibliography style to be 
%%% applied, and, second, the bibliography file to be used.

\bibliographystyle{ACM-Reference-Format} 
\bibliography{paper}

%%%%%%%%%%%%%%%%%%%%%%%%%%%%%%%%%%%%%%%%%%%%%%%%%%%%%%%%%%%%%%%%%%%%%%%%

\newpage
\onecolumn

\begin{appendix}

\section{Playing Games with Adversarial Bandits} 
\label{app:games-adv-bandits}

Bandit algorithms quantify their measure of error using regret, $R^T$, as the expected difference between the utility an algorithm achieves, and the utility it could have achieved over $T$ time steps had it chosen the best single arm in hindsight. Regret minimization algorithms aim to achieve sublinear regret so that $\lim_{n \rightarrow \infty} \frac{R^T}{T} = 0$.
Stochastic bandit algorihtms, such as UCB, achieve this by making stochastic assumptions on the source of the utility, \ie that they are generated by some underlying fixed distribution.
Adversarial bandits, on the other hand, allow an adversary to pick the distribution over utilities at each timestep. There are several algorithms that can achieve low regret even in this settings: Exp3~\cite{Auer98Exp3}, regret-matching~\cite{Hart01}, and generalized infinitesimal gradient ascent~\cite{Zinkevich03} are three examples.
There are well-established links to Nash equilibria in the case of two-player zero-sum games~\cite{Blum07}: the {\it average strategies} of two adversarial bandits in self-play converge to an approximate Nash equilibrium with high probability, with the approximation reducing to 0 over time.
Regret-matching, in particular, played an important role in scaling Poker AI (through counterfactual regret minmization~\cite{CFR}) and was also used in Diplomacy at large scale~\cite{Gray20Humanlevel}.

The setup is as follows. There is an unknown two-player zero-sum matrix game with row player strategies $S_1$ and column player strategies $S_2$ with payoff matrix $\mathbf{A}$ (utilities to the first player). The row player can adopt any mixed strategy in the simplex (probability distribution over $S_1$) $\bx \in \Delta(S_1)$ and similarly for column player ($\by \in \Delta(S_2)$). The goal is to find the minimax-optimal solution:
\[
\max_{\bx \in \Delta(S_1)} \min_{\by \in \Delta(S_2)} u_1(\bx, \by),
\]
where $u_1(\bx, \by) = \bx^{\top} \mathbf{A} \by$ is the expected value to the first player. Due to the Minimax theorem, this happens to coincide with the second player's optimization problem, so two two strategies that achieve the optimum are minimax-optimal (Nash) equilibrium.

Adversarial bandits converge to a minimax-optimal strategy over time in an iterative way via self-play. On each time step $t$, they employ a behavior strategies, \eg $\sigma^t_1$ and $\sigma^t_2$, and actions are sampled: $a_1^t \sim \sigma^t_1$, and $a_2^t \sim \sigma^t_2$.
They then observe a sampled utility $u_1(a_1^t, a_2^t)$  update internal quantities and choose a new strategy $\sigma^{t+1}_i$ based on these internal quantities. Since each one guarantees sublinear regret, the {\bf average strategies} $\bar{\sigma}^T$ converge to a minimax-optimal equilibrium.

We use regret-matching as described in~\cite[Section 4.4.3]{Bosansky16Algorithms}. 
The empirical matrix game is estimated by keeping track of the number of samples per $(a_i, a_j)$ pair, where $a_i \in S_1$ and $a_j \in S_2$ and total utility achieved when choosing $(a_i, a_j)$ such that there is a mean reward $\bar{u}^t_1(a_i, a_j)$.
On each iteration, each player has a current strategy $\sigma_i^t$, and selects actions on each round according to a sampling strategy: $a_i^t \sim \hat{\sigma}t_i(a) = \frac{\gamma}{|S|} + (1 - \gamma) \sigma_i^t$ to ensure adequate exploration.
The update is achieved by tracking cumulative regret for each action $R^t(a)$, and then applying:
\[
\sigma_i^{t+1}(a) = \frac{R^{t,+}(a)}{\sum_{a \in S_i} R^{t,+}(a)} \mbox{ if } \sum_{a \in S_i} R^{t,+}(a) > 0 \mbox { or } \frac{1}{|S_i|} \mbox{ otherwise },
\]
where $x^+ = \max(x, 0)$.
Let $u_1^t$ be the utility that was sampled from actions $(a_i^t, a_j^t)$ on step $t$, and $\textsc{Util}(b_1, b_2) = u_1^t$ if $(b_1, b_2) = (a_1, a_2)$ or $\bar{u}_1(b_1, b_2)$ otherwise.
The updates to the regrets of each player at time step $t$ are as follows:
\[
\forall b_1 \in S_1: R_1^{t+1}(b_1) = R_1^t(b_1) + (\textsc{Util}(b_1, a_2^t) - u_1^t)
\]
\[
\forall b_2 \in S_2: R_2^{t+1}(b_2) = R_2^t(b_2) - (\textsc{Util}(a_1^t, b_2) - u_1^t).
\]

\section{Variation in Task Rankings versus Ground Truth}
\label{app:data-variation}

How does one quantify the relationship between the ground truth, $\succ^*$ and the individual task rankings $\succ^*_v$?

\begin{figure*}[t!]
\begin{tabular}{cc}
\includegraphics[width=0.4\textwidth]{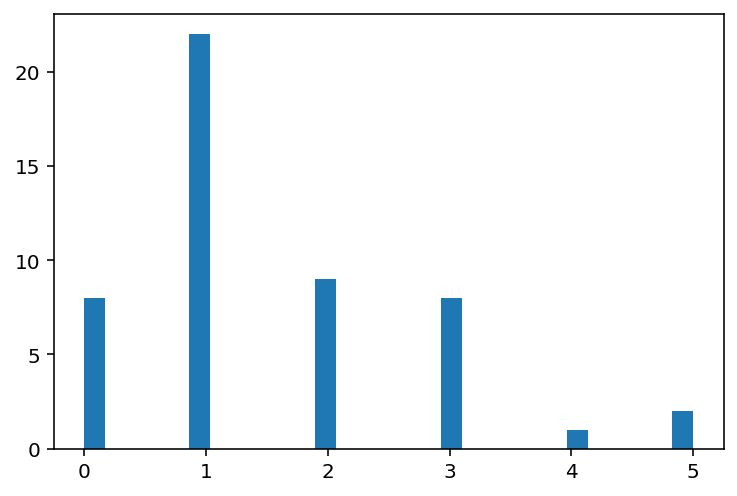} &
\includegraphics[width=0.4\textwidth]{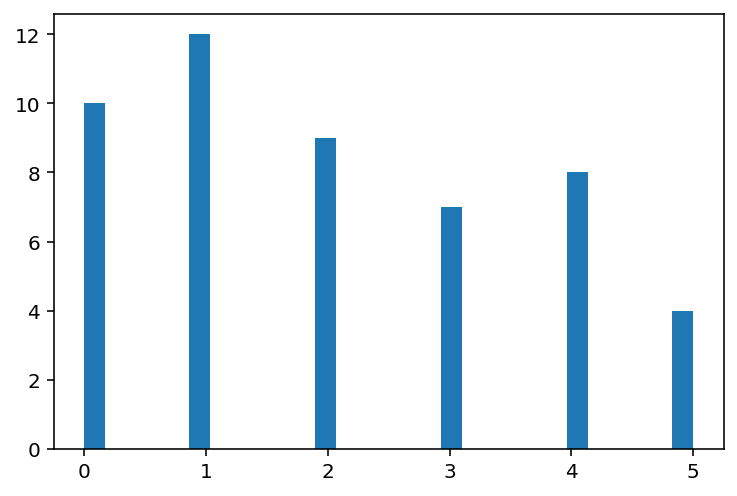} \\
$\phi = 0.2$ & $\phi = 0.3$ \\
\includegraphics[width=0.4\textwidth]
{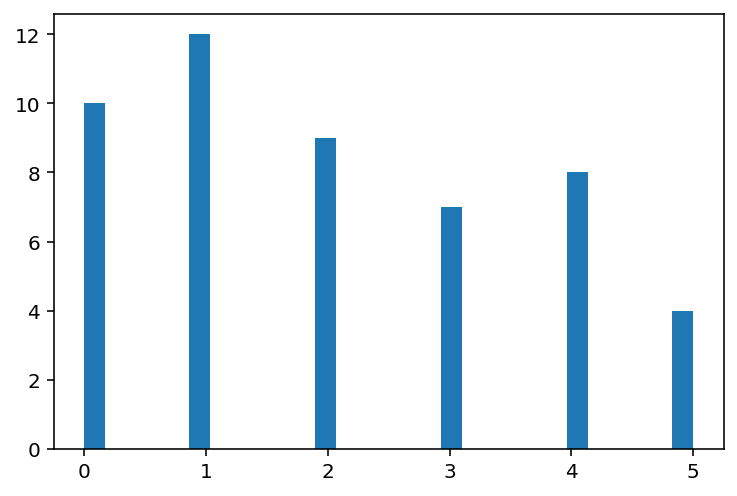} &
\includegraphics[width=0.4\textwidth]{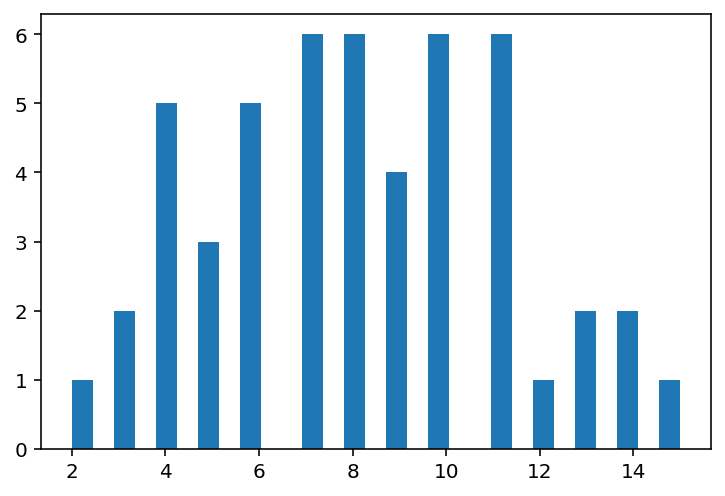} \\
$\phi = 0.5$ & $\phi = 0.6$ \\
\end{tabular}
\caption{Task variation for samples from a Mallows model with $m = 8, n = 50$. Each graph shows a histogram of the 50 Kendall-tau distances between the task ranking and the ground truth ranking, $K_d(\succ^*, \succ^*_v)$. }
\label{fig:synthetic-data-variation}
\end{figure*}

To give a very rough depiction , Figure~\ref{fig:synthetic-data-variation} shows a sample histogram of the Kendall-tau distances $K_d(\succ^*, \succ^*_v)$ for the Mallows model with $m=8$ agents and $n=50$ tasks, for various values of $\phi$.

\begin{figure*}[t!]
\begin{center}
\includegraphics[width=0.4\textwidth]{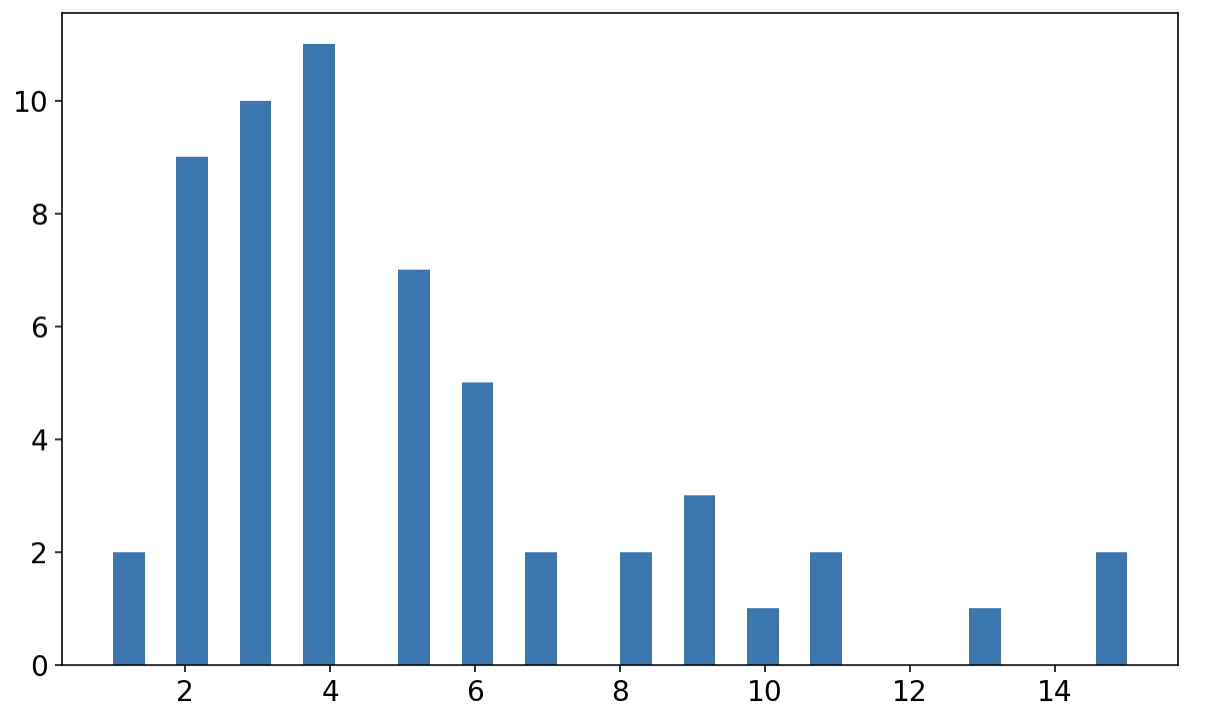}
\end{center}
\caption{Task variation for Atari Agent 57 data generator with $m = 8, n = 57$.}
\label{fig:atari-data-variation}
\end{figure*}

For comparison, the task variation in the Atari Agent57 data set is shown in Figure~\ref{fig:atari-data-variation}.


\subsection{Kemeny as an Approximation to the Ground Truth}
\label{app:kemeny-ground-truth}

How hard is it to find the ground truth ranking from the generated task rankings $\{ \succ_v^* \}_{v \in V}$ or from a set of score samples? In this section, we analyze the difficulty of the problem on the Mallows models used in the paper.

First, we must define an optimal ranking under complete ranking data. Given a set of complete ranked-ballot votes $V$ (each a total orders over $A$), which we shall refer to as $\{ \succ_v \}_{v \in V}$. We define the {\bf optimal ranking under the data}~\cite[Definition 3]{Lanctot25SCO} as the ranking that minimizes the average Kendall-tau distance to all the votes:
\begin{equation}
\hat{\succ}^* = \min_{\succ'} \frac{1}{|V|} K_d(\succ', \succ_v).
\label{eq:opt-ktd-ranking}
\end{equation}

There is a voting method that finds such a ranking: the Kemeny voting method~\cite{Kemeny59}. It is computationally hard to compute and cannot be used in practice with a high number of agents, so it is impractical when $|A| = m$ is larger than 20, a problem for active evaluation in general. However, it can be used in our case to check the difficulty because $m = 8$.

Define the preference function $N(a_i, a_j)$ as the number of votes in $V$ where $a_i$ is ranked higher than $a_j$. The Kemeny score of a ranking is:
\[
\textsc{KemenyScore}(\succ) = \sum_{a_i~\succ~a_j} N(a_i, a_j).
\]
For example, given a preference profile with $|A| = m = 3$ and preference function $N$, the ranking:
\[
a_2 \succ a_0 \succ a_1,
\]
would have a Kemeny score of $N(a_2, a_0) + N(a_2, a_1) + N(a_0, a_1)$. The Kemeny voting method returns a ranking with a maximal Kemeny score: $\argmax_{\succ'} \textsc{KemenyScore}(\succ')$.

Equipped with these tools, we can now return to answer the question of difficulty. Before addressing the practical question, note that the Kemeny ranking is a maximum likelihood estimator of the ground truth under the Mallows model~\cite[Chapter 8]{Brandt16Handbook}, so it is a natural choice for the ground truth ranking given some evaluation data. But can it recover the ground truth in practice?

\subsubsection{Ground Truth from Task Rankings}
\label{sec:kemeny-gt-from-task-rankings}

As a Kemeny ranking (solution to Equation~\ref{eq:opt-ktd-ranking}) is not unique, we also use Kemeny score distance (KSD) to indicate proximity to the ground truth; \ie given a ground truth ranking from a Mallows data generator $\succ^*$ and votes $V$ from task rankings $\succ_v^*$ described in Section~\ref{sec:synthetic-data},
\[
\textsc{KemenyScoreDistance}(\succ^*, succ') = |\textsc{KemenyScore}(\succ^*) - \textsc{KemenyScore}(\succ')|.
\]
If the Kemeny score distance for some ranking $\succ'$ is 0, then it has the same Kemeny score as the ground truth, and is a solution to Equation~\ref{eq:opt-ktd-ranking}.

We generate 1000 seperate instances of Mallows models, each with with $m = 8$ and $n = 50$ using dispersion $\phi \in \{0.3, 0.6\}$. 
Each instance $t \in \{1, \cdots, 1000 \}$ is represented by task rankings $V^t$, where  $|V^t| = n = 50$ votes, and the Kemeny ranking $\succ^t$ is computed using $V^t$
The results are shown in Table~\ref{tab:kemeny-gt-test-task-rankings}. 
\begin{table}
\begin{tabular}{c|ccc|}
$\phi$ & $\bar{K}_n(\succ^*, \succ^t)$ & Average $\textsc{KSD}(\succ^*, \succ^t)$ & Num of instances that achieve 0 KSD \\
\hline
0.3 & 0     & 0     & 1000 \\
0.6 & 0.005 & 0.692 &  822 \\
\hline
\end{tabular}
\caption{Kemeny rankings proximity to the ground truth.}
\label{tab:kemeny-gt-test-task-rankings}
\end{table}
When $\phi = 0.3$, the Kemeny ranking on every instance finds the ground truth ranking. When $\phi = 0.6$, there is some error due to the variation, but the approximation is very close: the average normalized KtD is less than $0.005$, average Kemeny score distance is less than a single vote, and 82.2\% of instances achieve 0 KSD. We conclude from this that a ranking that satisfies Equation~\ref{eq:opt-ktd-ranking} is a very good approximation of the ground truth, in practice.

\subsubsection{Ground Truth from Score Distributions}

\begin{figure*}[t]
\begin{tabular}{cc}
\includegraphics[width=0.48\textwidth]{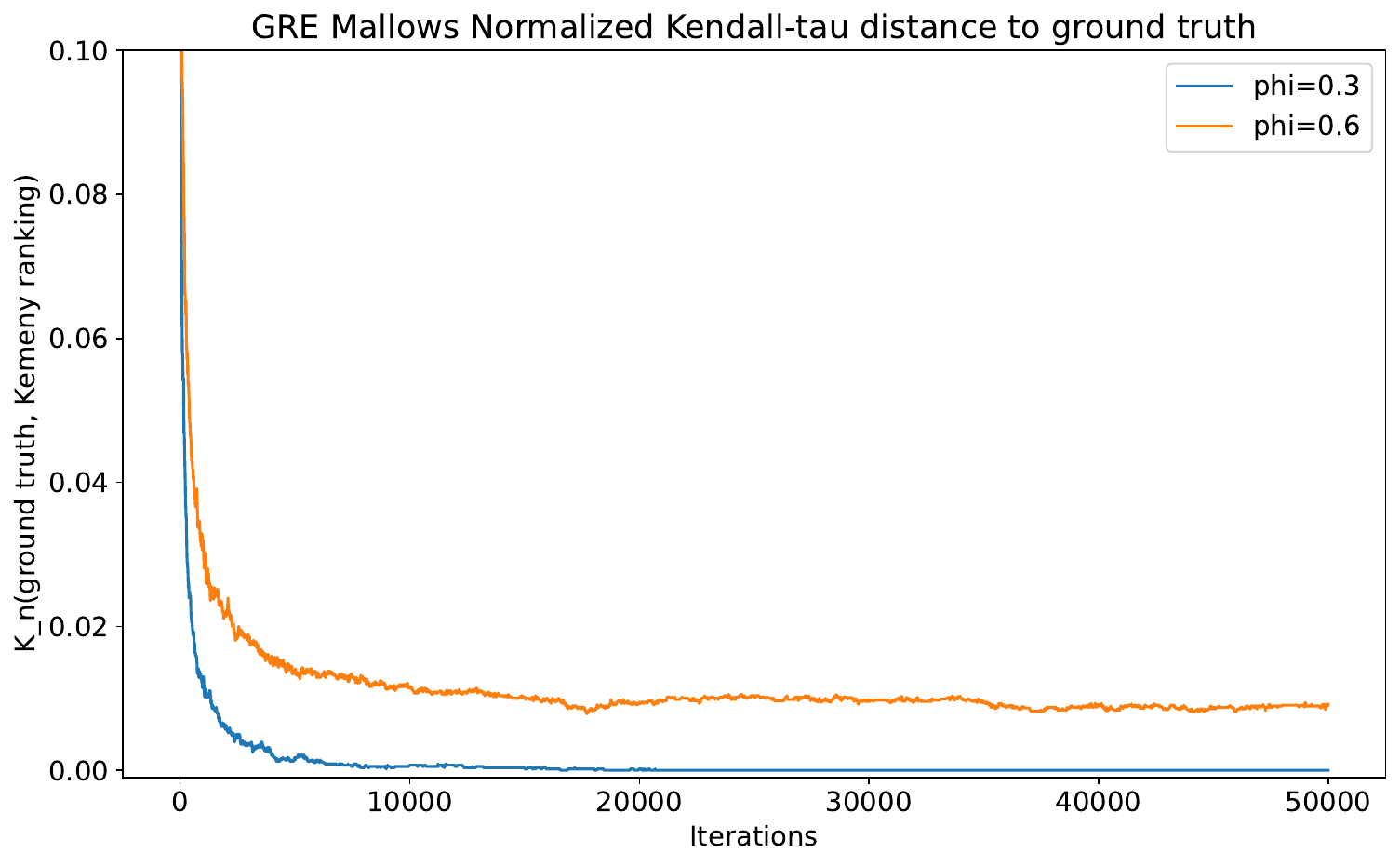} &
\includegraphics[width=0.48\textwidth]{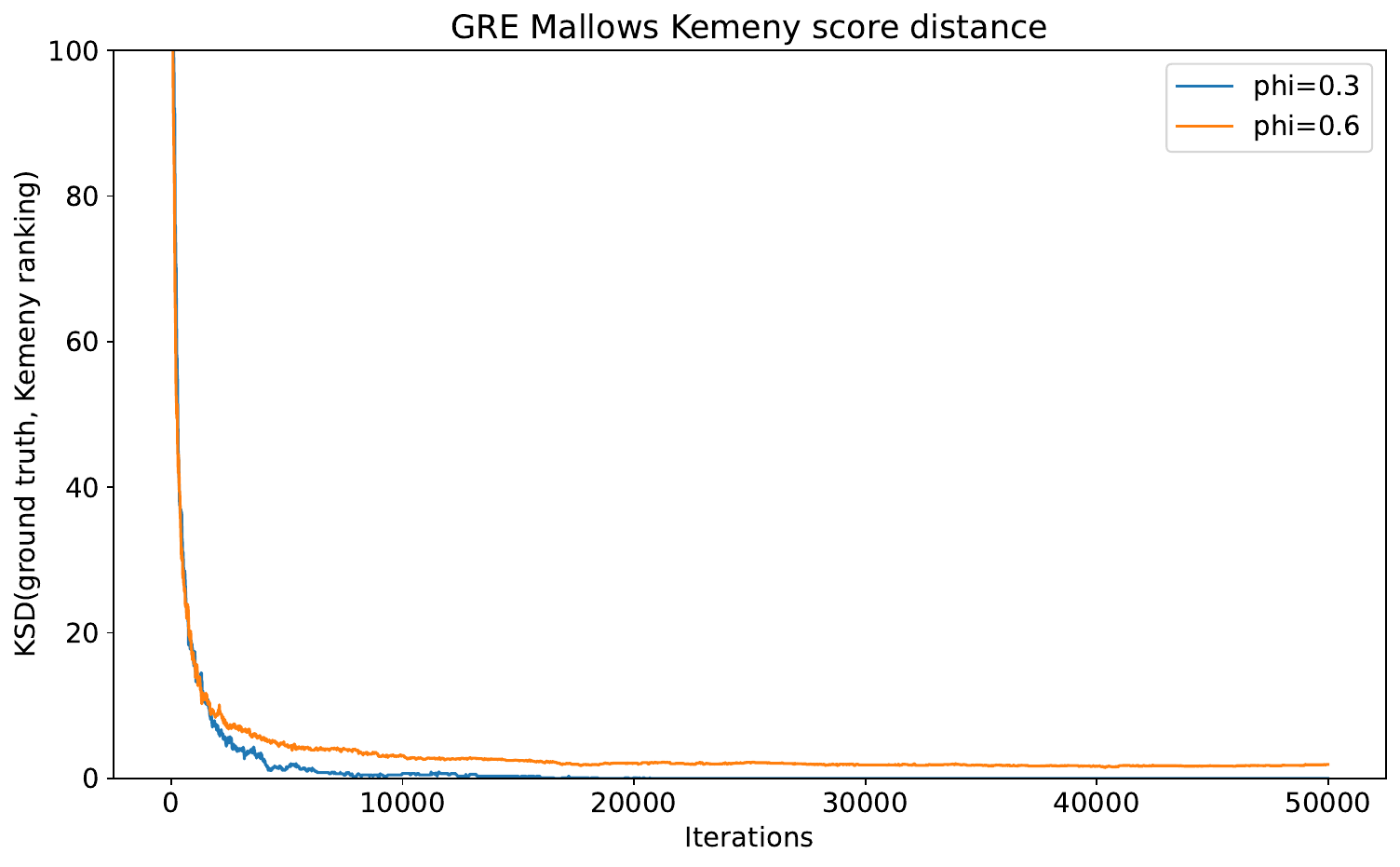} \\
\end{tabular}
\caption{}
\label{fig:kemeny-gt-test-score-dist}
\end{figure*}

We were also curious how difficult the problem becomes in our setting where feedback to algorithms are noisy. So, we also run sampled scores from the underlying score comparisons from the underlying score distributions $S_{v,a}$ randomly selecting tasks and agents on every iteration, and reporting the Kemeny ranking $\succ^t$ on interation $t$ on the growing data set as the algorithms do in Algorithm~\ref{alg:active-eval-loop}. The results are shown in Figure~\ref{fig:kemeny-gt-test-score-dist}.

Similarly to the Section~\ref{sec:kemeny-gt-from-task-rankings}, every instance eventually finds the ground truth ranking under $\phi = 0.3$ yielding normalized Kendall-tau and Kemeny score distance of 0. When $\phi = 0.6$, the average normalized Kendall-tau distance is approximately $0.01$ and average Kemeny score distance is $1.92$.

\section{Additional Results}
\label{app:additional-results}

\subsection{Comparison of Group Sizes in Proportional Representation}
\label{sec:proprep-group-sizes}

\subsubsection{Synthetic Data using Mallows Data Generator}

\begin{table}
\begin{center}
\begin{tabular}{c|cc}
{\bf Group fraction} ($g_f$) & {\bf Average Task Subset size} $|V^t|$ & {\b Percentage Tasks Reduction} (\%) \\
\hline
0.10  & 50.00 & 0   \\
0.20  & 47.11 & 5.78 \\
0.25 &  47.11 & 5.78 \\
0.30 &  38.09 & 23.82 \\
0.35 &  38.07 & 23.86 \\
0.50 &  30.85 & 38.30 \\
0.60 &  26.25 & 47.50      \\
0.75 &  21.75 & 56.50      \\
\end{tabular}   
\end{center}   
\caption{Effect on task set reduction based on group fraction parameter $g_f$. Average size varied very little over time $t$, and was consistent across experiments ($(\phi, k) in \{0.3, 0.6\} \times \{3, 8\}$ so we only show the overall average.}
\label{tab:proprep_group_sizes}
\end{table}

\begin{figure*}[t]
\begin{tabular}{cc}
\includegraphics[width=0.48\textwidth]{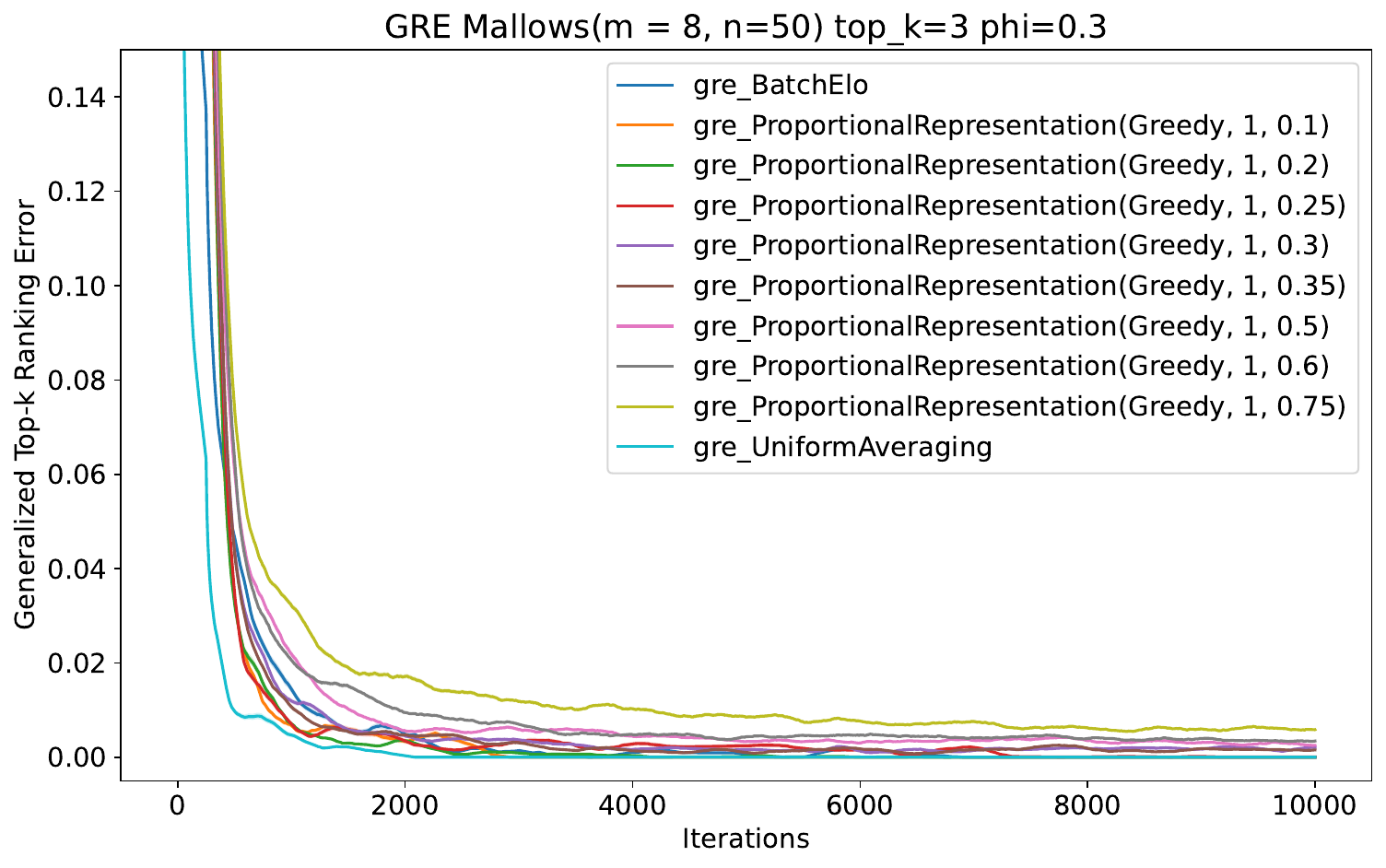} &
\includegraphics[width=0.48\textwidth]{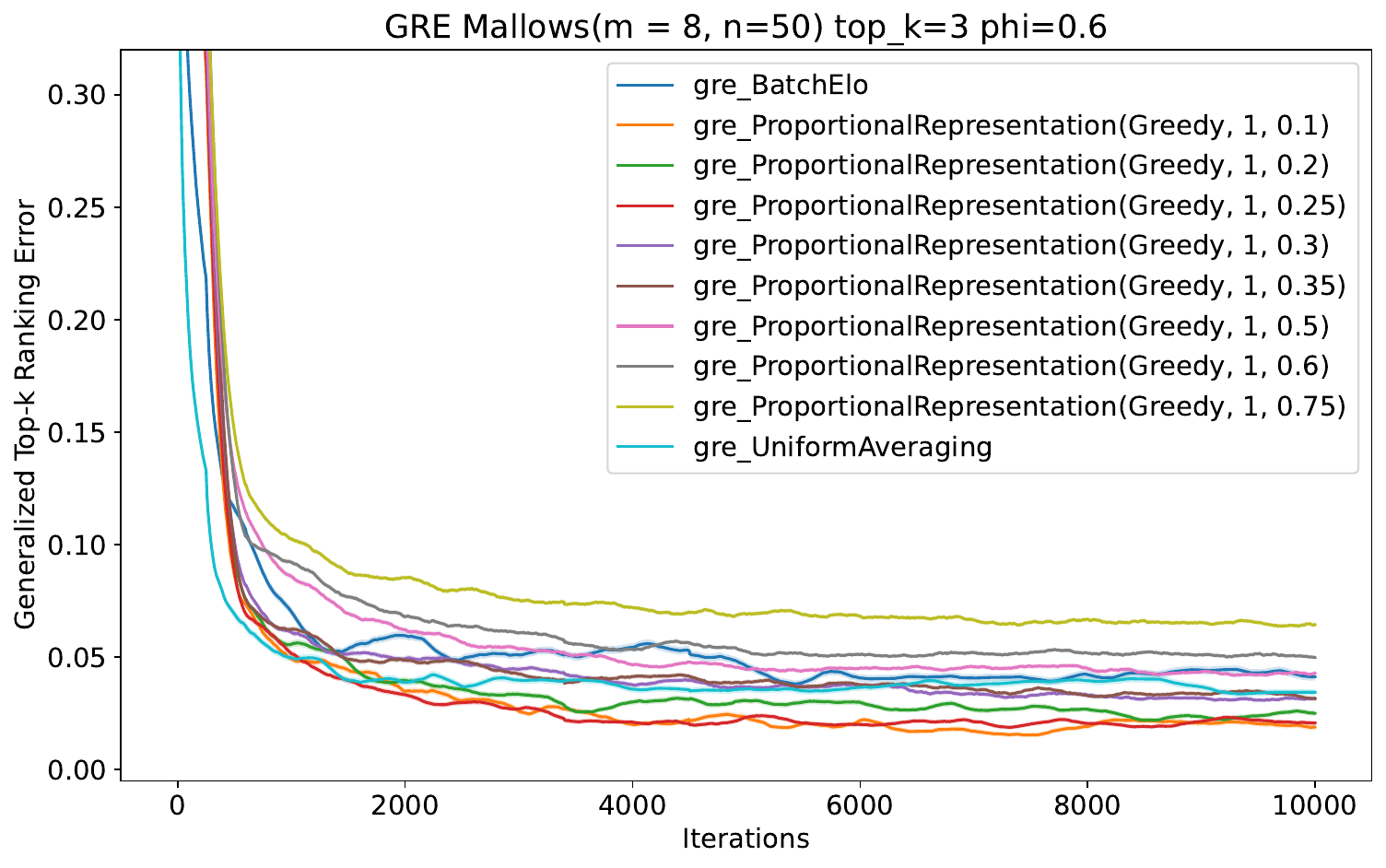} \\
\includegraphics[width=0.48\textwidth]{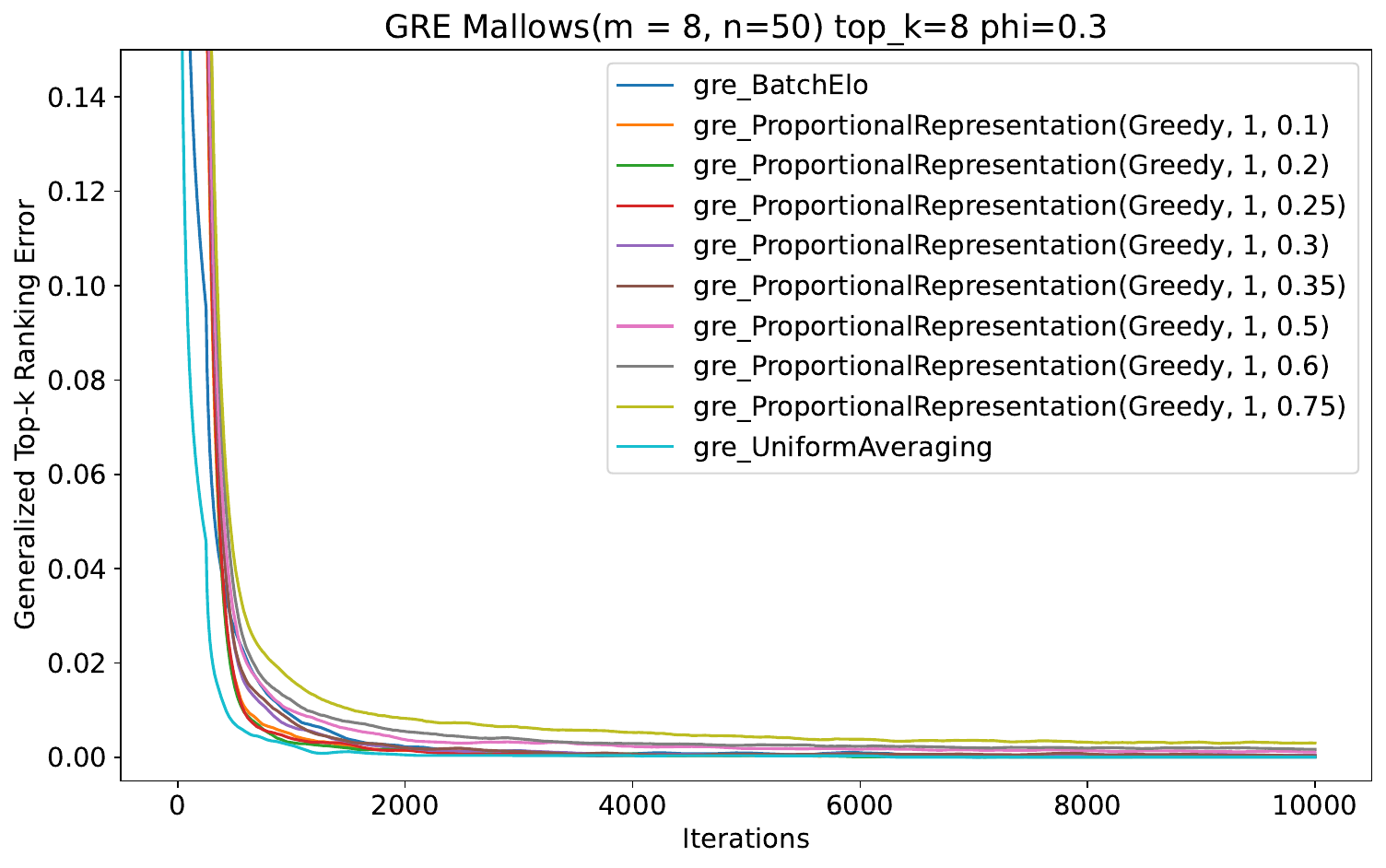} &
\includegraphics[width=0.48\textwidth]{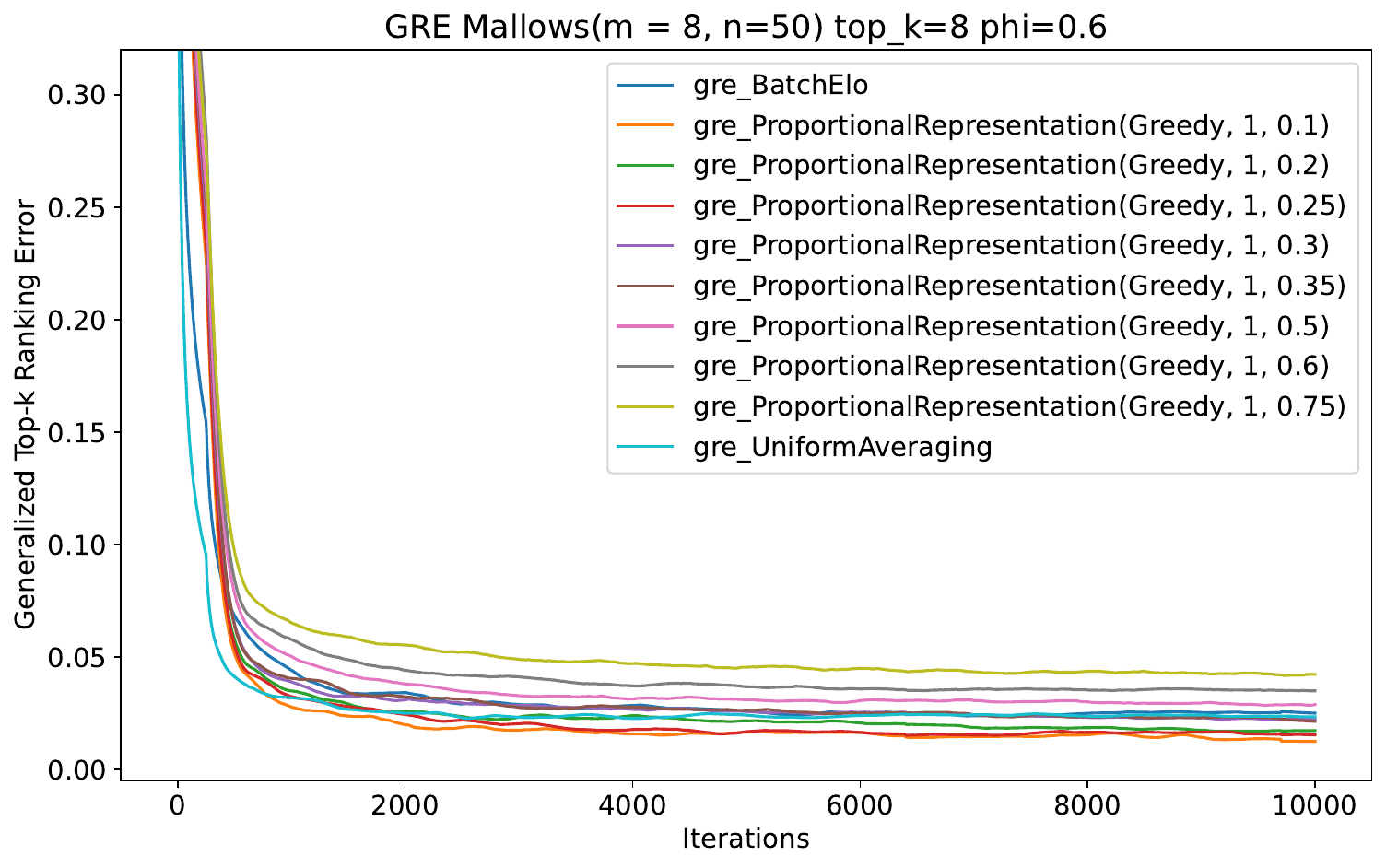} \\
\end{tabular}
\caption{GRE over time for the \textsc{ProportionalRepresentation} algorithm on the Mallows model for different settings of $g_f \in \{0.1, 0.2, 0.25, 0.3, 0.35, 0.5, 0.6, 0.75\}$ and $(\phi, k) \in \{0.3, 0.6\} \times \{3, 8\}$.}
\label{fig:proprep_group_sizes}
\end{figure*}

In this subsection, we analyze the effect of the group size fraction parameter ($g_f$) on the \textsc{ProportionalRepresentation} algorithm on the Mallows synthetic data model, with $m = 8$ agents and $n = 50$ tasks (Experimental condition 1).

For any given run, the algorithm uses a reduced task set $V^t \subseteq V$ on each iteration.
We define the group size $g = \lceil g_f n \rceil$ where $g_f \in [0,1]$ is a fraction of $n$.
Table~\ref{tab:proprep_group_sizes} and Figure~\ref{fig:proprep_group_sizes} shows the effect of various choices of $g_f$ on the GRE and empirical sizes of the reduced task sets.
We observe that in the case of $\phi = 0.3$, all of the algorithms reduce GRE quite quickly showing little difference, except $g_f = 0.75$ is clearly inferior.
However, when $\phi = 0.6$ all settings $g_f \in \{0.1, 0.2, 0.25, 0.3, 0.35\}$ either match the best baseline or outperform it in the long run, while reducing the task sets by nearly a quarter.
In the end, $g_f = 0.1$ performs best and retains all the tasks; in this case, the source of the benefit could be the mean-model aggregation when reporting the final ranking.

\subsubsection{Atari Data}

We ran a similar experiment to show the effect of $g_f$ on the reduced task sets and GRE over time. Results are shows in Tables~\ref{tab:proprep_group_sizes_atari1} and \ref{tab:proprep_group_sizes_atari2}. The results and observations are quite consistent with the synthetic data set.

\begin{table}
\begin{center}
\begin{tabular}{c|cc}
{\bf Group fraction} ($g_f$) & {\bf Average Task Subset size} $|V^t|$ & {\b Percentage Tasks Reduction} (\%) \\
\hline
0.10  & 57.00 & 0   \\
0.20  & 53.45 & 6.23 \\
0.25 &  53.43 & 6.26 \\
0.30 &  44.29 & 22.3 \\
0.35 &  44.27 & 22.33 \\
0.50 &  35.70 & 37.37 \\
0.60 &  29.34 & 48.53  \\
0.75 &  24.65 & 56.75  \\
\end{tabular}   
\end{center}   
\caption{Effect on task set reduction based on group fraction parameter $g_f$ in the Atari data set. Average size varied very little over time $t$, and was consistent across experiments.}
\label{tab:proprep_group_sizes_atari1}
\end{table}

\begin{table}
\begin{center}
\begin{tabular}{c|cc}
{\bf Group fraction} ($g_f$) & AGRE($k = 3$) & AGRE($k = 8)$ \\
\hline
0.1   & 0.045472 & 0.013558 \\
0.20  & 0.050451 & 0.015431 \\
0.25  & 0.056001 & 0.016220 \\
0.30  & 0.074925 & 0.022702 \\
0.35  & 0.079511 & 0.023969  \\
0.50  & 0.096508 & 0.031637  \\
0.60  & 0.109969 & 0.038448  \\
0.75  & 0.122674 & 0.044319  \\
\end{tabular}   
\end{center}   
\caption{Effect on task set reduction based on group fraction parameter $g_f$ in the Atari data set.}
\label{tab:proprep_group_sizes_atari2}
\end{table}

% 0.1 57.0
% 0.2 53.447212
% 0.25 53.429169
% 0.3 44.291594
% 0.35 44.27487
% 0.5 35.700252
% 0.6 29.341696
% 0.75 24.657726

% 0.1 57.0
% 0.2 53.447212
% 0.25 53.429169
% 0.3 44.291594
% 0.35 44.27487
% 0.5 35.700252
% 0.6 29.341696
% 0.75 24.657726

\subsection{Robustness to Clones}
\label{sec:expcond3-clones}

We now assess the robustness of each algorithm to the presence of cloned agents. 
Clone-invariance is an important motivation behind the development of game-theoretic ratings~\cite{Balduzzi18ReEval,Marris25Deviation} and Elo can be susceptible to clones~\citep{lanctot2023evaluating,liu2025reevaluatingopenendedevaluationlarge}. Ranked Pairs is clone-consistent, ensuring that the top-ranked agent would not change in the presence of clones in the classical voting setting.
We assess the effects of clones in active evaluation.
Specifically, we repeat Experimental Condition 1 with the following modification to the data generator. When one of the $m=8$ agents is {\it cloned}, say $a'$ is a clone of $a$, a coin is flipped. If the coin lands heads, for each task ranking $\succ_v^*$: $a'$ is inserted directly ahead of $a$ (and vice versa if coin lands tails: $a'$ is inserted directly behind $a'$ in the ranking), and similarly for the ground truth ranking. The mean of the score distribution for $a'$ is then a slight perturbation: $\mu_{v,a'} = \mu_a + \epsilon (\mu_{v,b} - \mu_{v,a})$, where $b$ is the agent directly above/below $a$ in $\succ_v^*$ (direction depending on the coin flip).

For each run we add 8 clones (randomly chosen, one by one, using the method above) of the original 8 agents so that $m=16$.
We then run all of the algorithms on the modified data generator, at each step extracting assessing the ranking among only the original 8 agents.
Results are shown in Table~\ref{tab:expcond3-clones-agre}.
First, comparing to Table~\ref{tab:expcond1-tab-agre} we notice a general increase in absolute ranking error due to the introduction of clones, particularly for $k=3$. However, the trends remain consistent and the baseline algorithms still reduce the error rate fastest. Outside the \textsc{BasicUCB} and \textsc{UniformAveraging}, the mean-model variants and \textsc{BatchElo} perform best.

\begin{table}
\begin{tabular}{c|cc}
{\bf Algoirthm} & $\textsc{AGRE}(k=3)$ & $\textsc{AGRE}(k=8)$ \\
\hline
\textsc{BasicUCB}              & {\bf 0.004598} &  0.013604 \\
\textsc{UniformAveraging}      & 0.005189       &  {\bf 0.003433} \\
\textsc{MeanModelCopeland}     & 0.015629       &  0.010145          \\
\textsc{MeanModelMaxLot}       & 0.016105       &  0.010282          \\
\textsc{BatchElo}              & 0.016640       &  0.009298  \\
\textsc{MeanModelRankedPairs}  & 0.017065       &  0.010222            \\
\textsc{OnlineElo}             & 0.017744       &  0.010936  \\
\textsc{OnlineSCO}             & 0.018654	    &  0.012864  \\
\textsc{OnlineNashAvg}         & 0.019282       &  0.010248  \\
\textsc{BatchSCO}              & 0.021251       &  0.013602  \\
\textsc{ProportionalRepresentation}   & 0.032313  &  0.020639  \\
\textsc{BatchRankedPairs}      & 0.038718       &  0.025355  \\
\textsc{BatchCopeland}         & 0.039895       &  0.024524  \\
\textsc{BatchMaximalLotteries} & 0.054654       &  0.035792  \\
\textsc{OnlineMaxLot}          & 0.063327       &  0.072039  \\
\textsc{BatchNashAvg}          & 0.178460       &  0.116436  \\
\hline
\end{tabular}
\caption{Average Generalized Top-$k$ ranking error, $\textsc{AGRE}(\succ^{1:t}, \succ^*, t, k = \cdot)$ at $t = 10^4$ with 8 cloned agents. 95\% confidence intervals for each entry was $< 10^{-3}$.}
\label{tab:expcond3-clones-agre}
\end{table}

\subsection{Results using Plackett-Luce Model}
\label{app:plackett-luce}

The Plackett-Luce model is a generalization of the Bradley-Terry model which is the basis of the Elo rating system mentioned in Section~\ref{sec:eval-systems}~\cite{Plackett75,Luce59}.
As in Elo, each agent $a_i$ where $i \in \{ 1, 2, \cdots, m \}$ is assigned a 
rating (numerical value) $\theta_i$.
Ratings are generated uniformly at random in a bounded interval, and then sorted in descending order to obtain the ground truth $\succ^*$.
Then, each task ranking $\succ^*_v$ is generated in a way that is similar to drawing colored balls from an urn in sequence (without replacement): at first, the urn is full and contains $m$ balls, and balls (agents) are removed one-by-one until there are none left. The probability of sampling an agent on each step is
\[
Pr(a_i \sim A') = \textsc{Softmax}(a_i, \btheta, A', \tau) = \frac{\exp{( \theta_i / \tau)}}{\sum_{a_j \in A'} \exp{(\theta_j / \tau)}},
\]
where $A' \subseteq A$ is the set of agents remaining after previous removals and $\tau$ is a temperature that controls the added noise to the sampling process. The order the agents are selected determines the task ranking $\succ^*_v$ (first one is the top-ranked agent and last one is the bottom-ranked agent).
Then, task score distributions $S_{v,a}$ are generated in the same way as desfribed in Section~\ref{sec:mallows}.

\begin{figure*}[t]
\begin{tabular}{cc}
\includegraphics[width=0.48\textwidth]{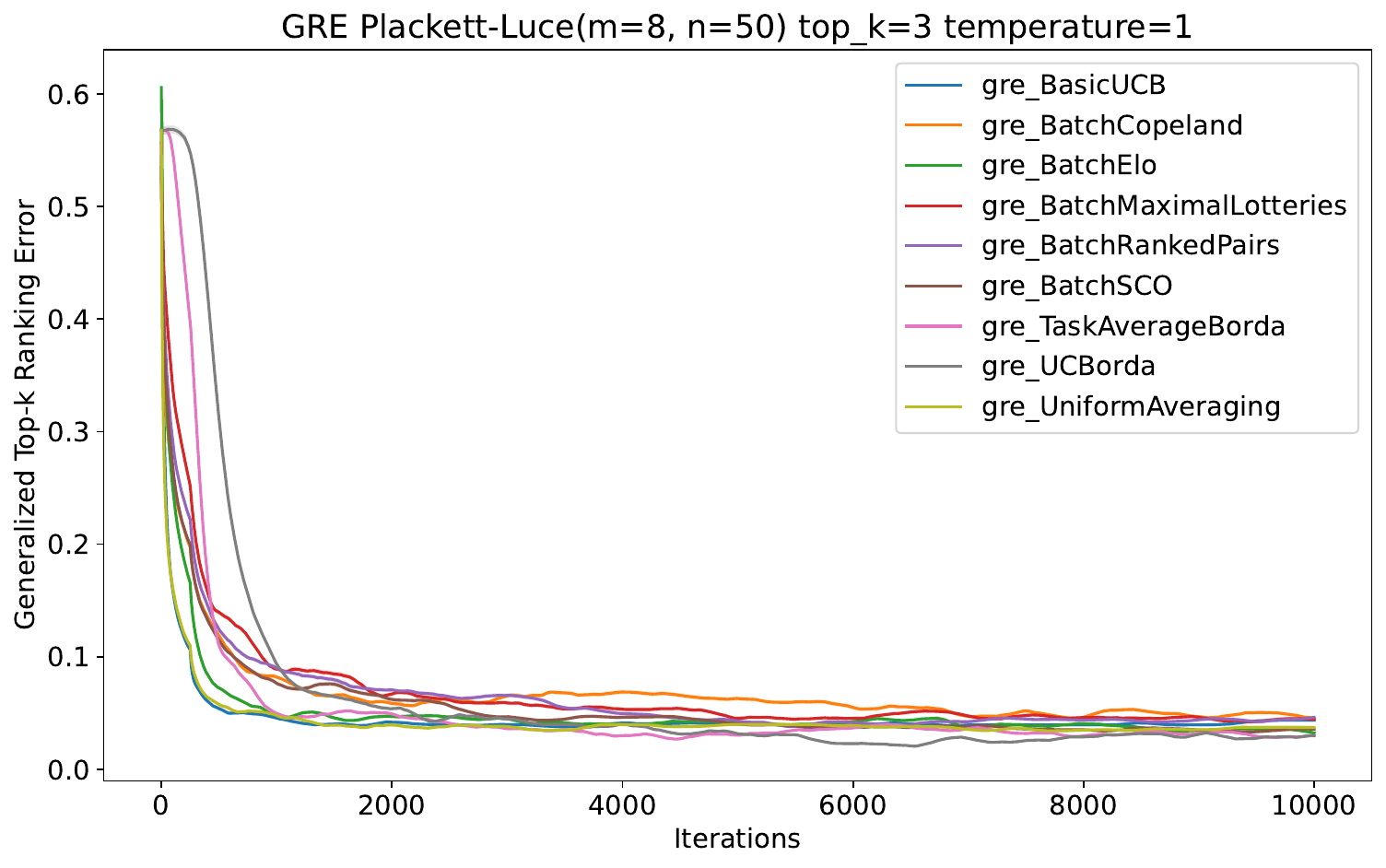} &
\includegraphics[width=0.48\textwidth]{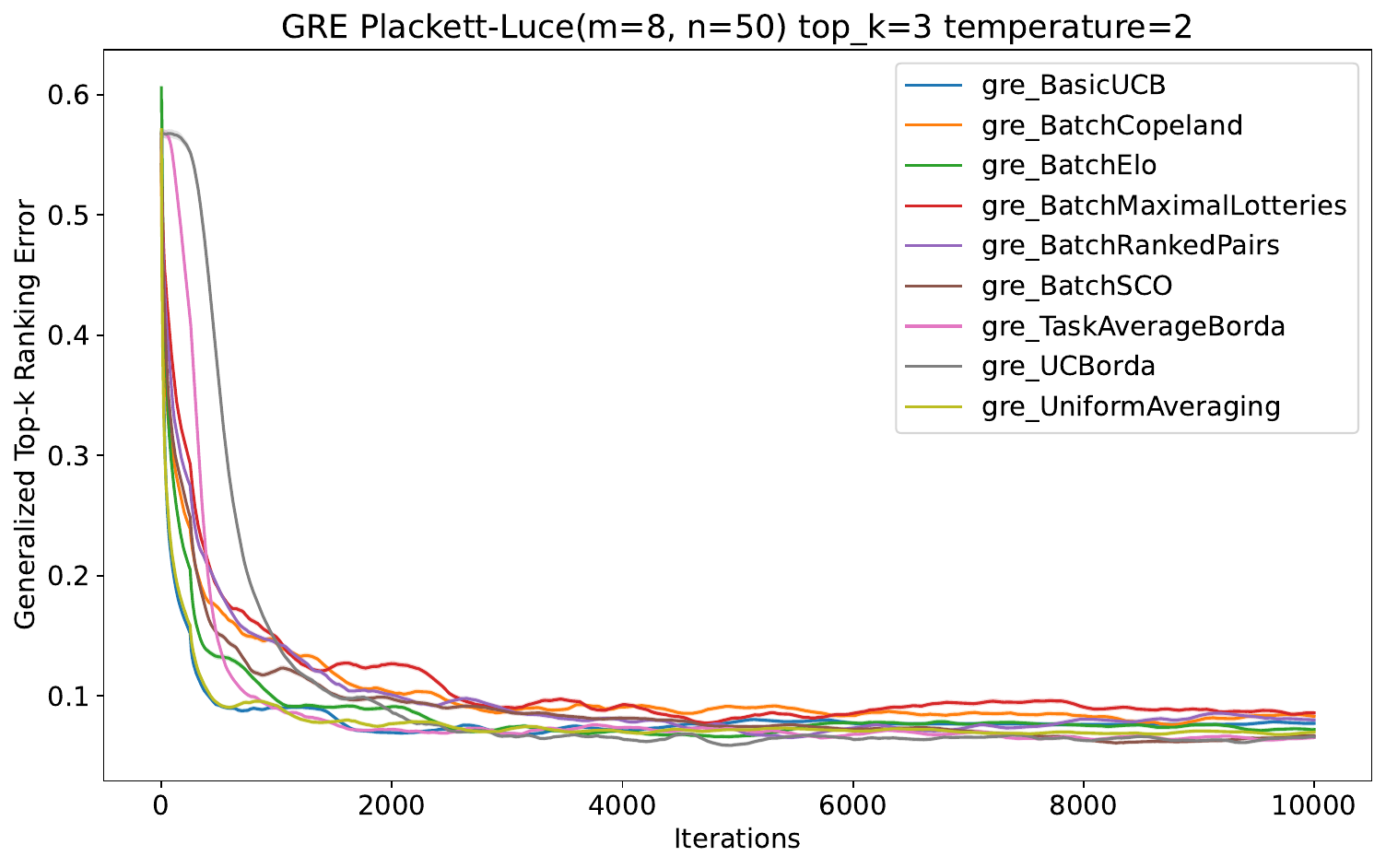} \\
\includegraphics[width=0.48\textwidth]{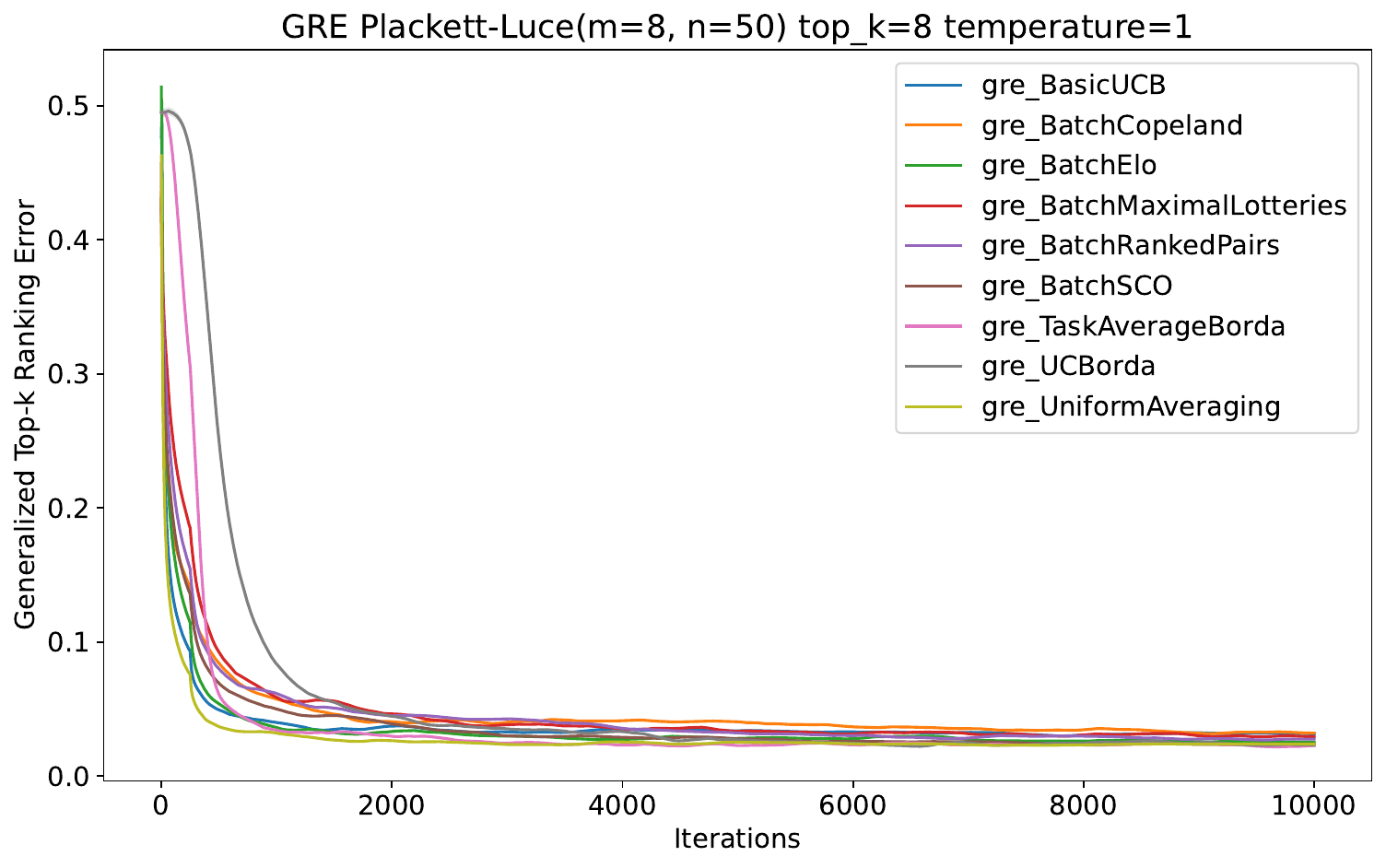} &
\includegraphics[width=0.48\textwidth]{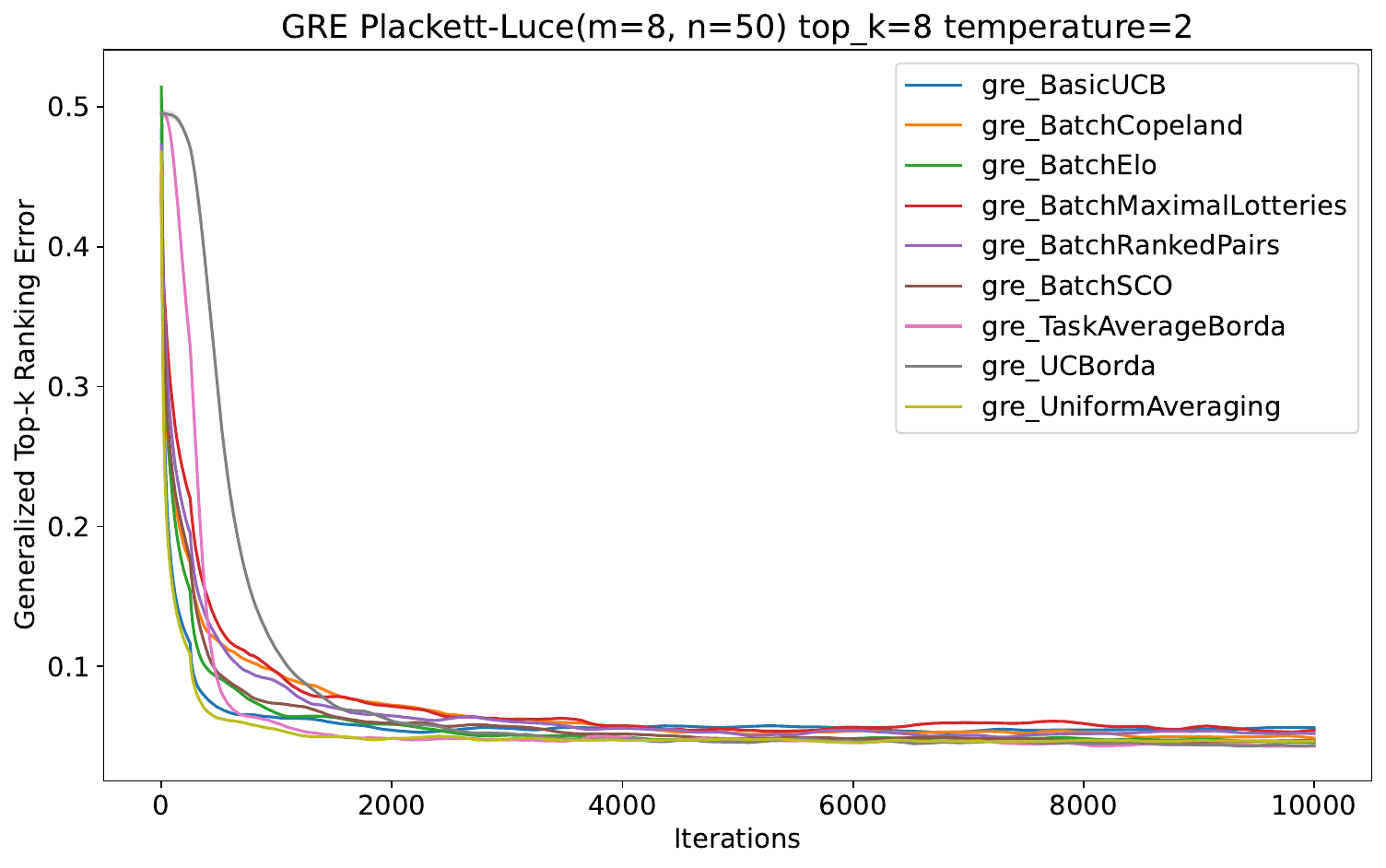} \\
\end{tabular}
\caption{Generalized Top-k Ranking Error on synthetic data generated by Plackett-Luce model set baseline \& batch algorithms, for varying temperatures $\tau \in \{1, 2\}$ (left versus right column) and varying $k \in \{3, 8\}$ (top vs. bottom row).}
\label{fig:expcond1-pl-gre-graphs}
\end{figure*}

\subsection{Atari Agent57 Results}

\begin{figure*}[t]
\begin{tabular}{cc}
\includegraphics[width=0.48\textwidth]{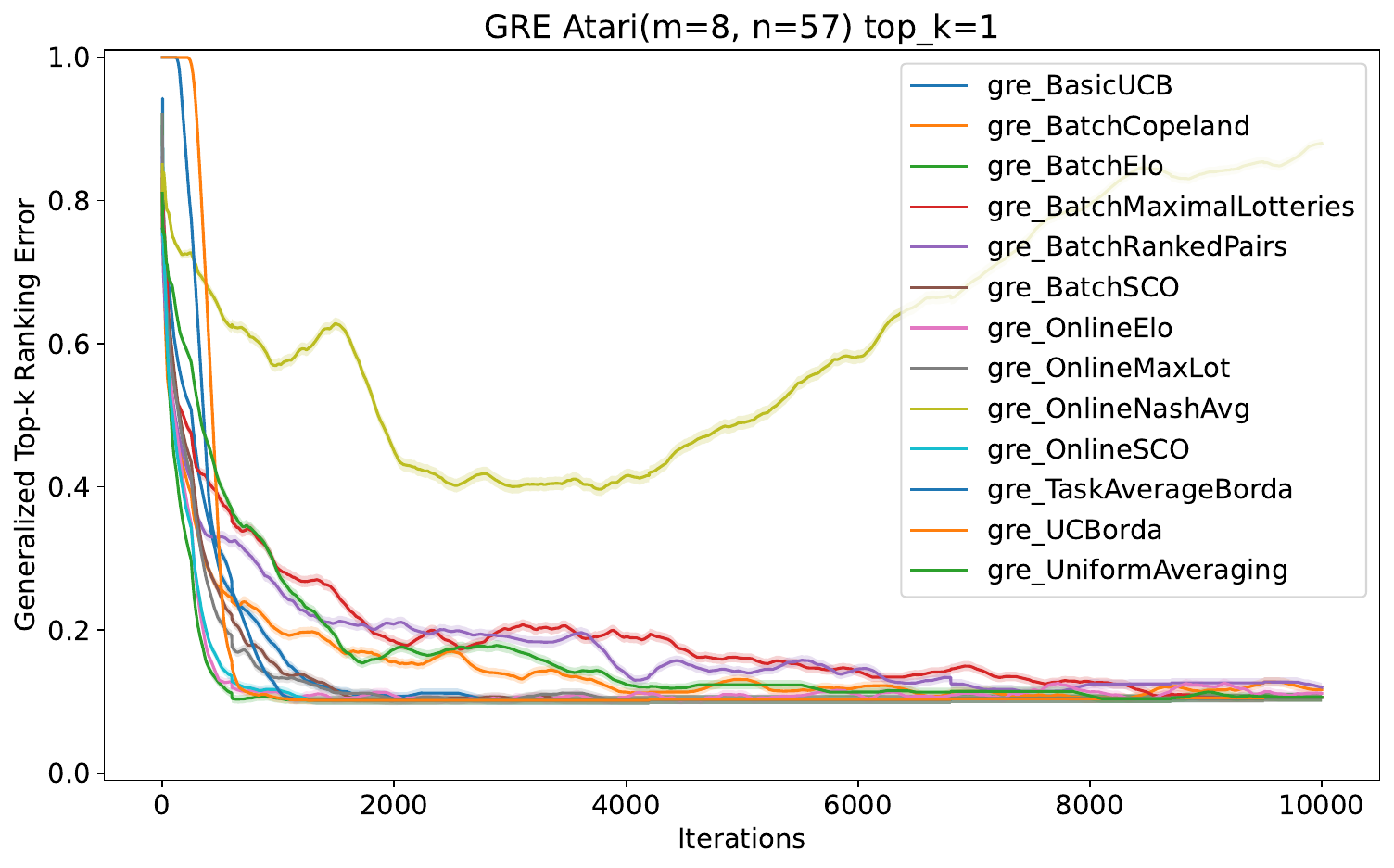} &
\includegraphics[width=0.48\textwidth]{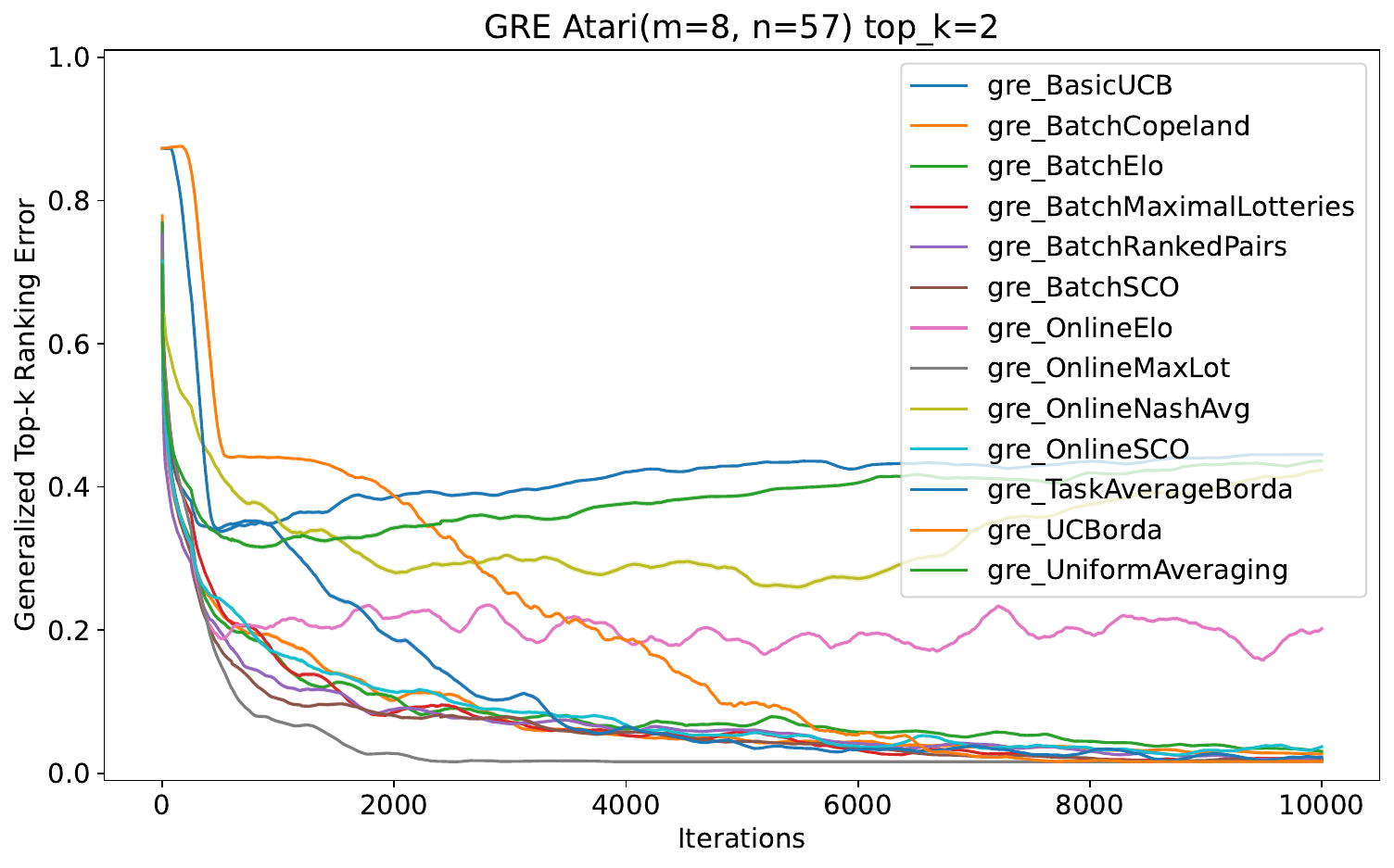} \\
\includegraphics[width=0.48\textwidth]{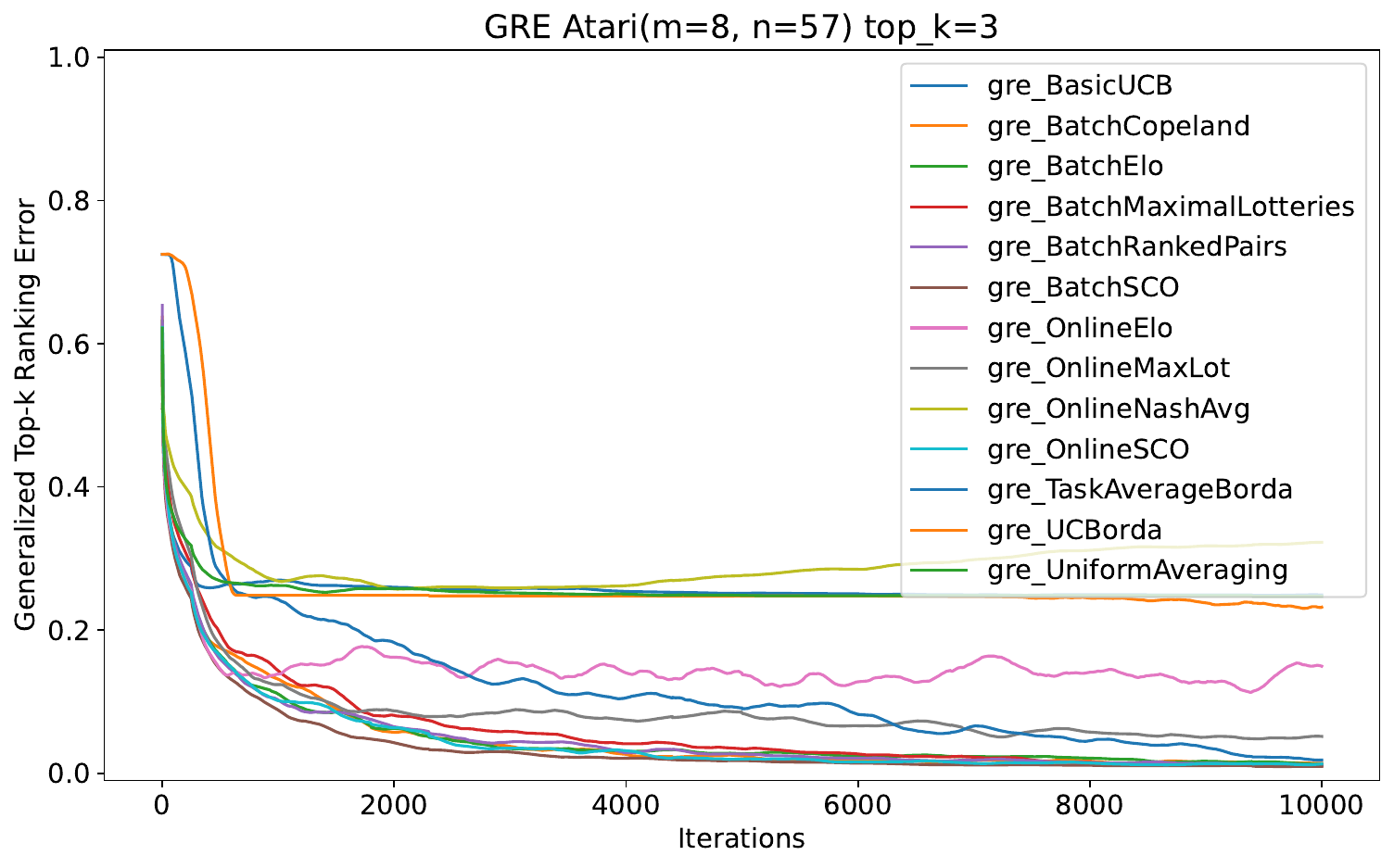} &
\includegraphics[width=0.48\textwidth]{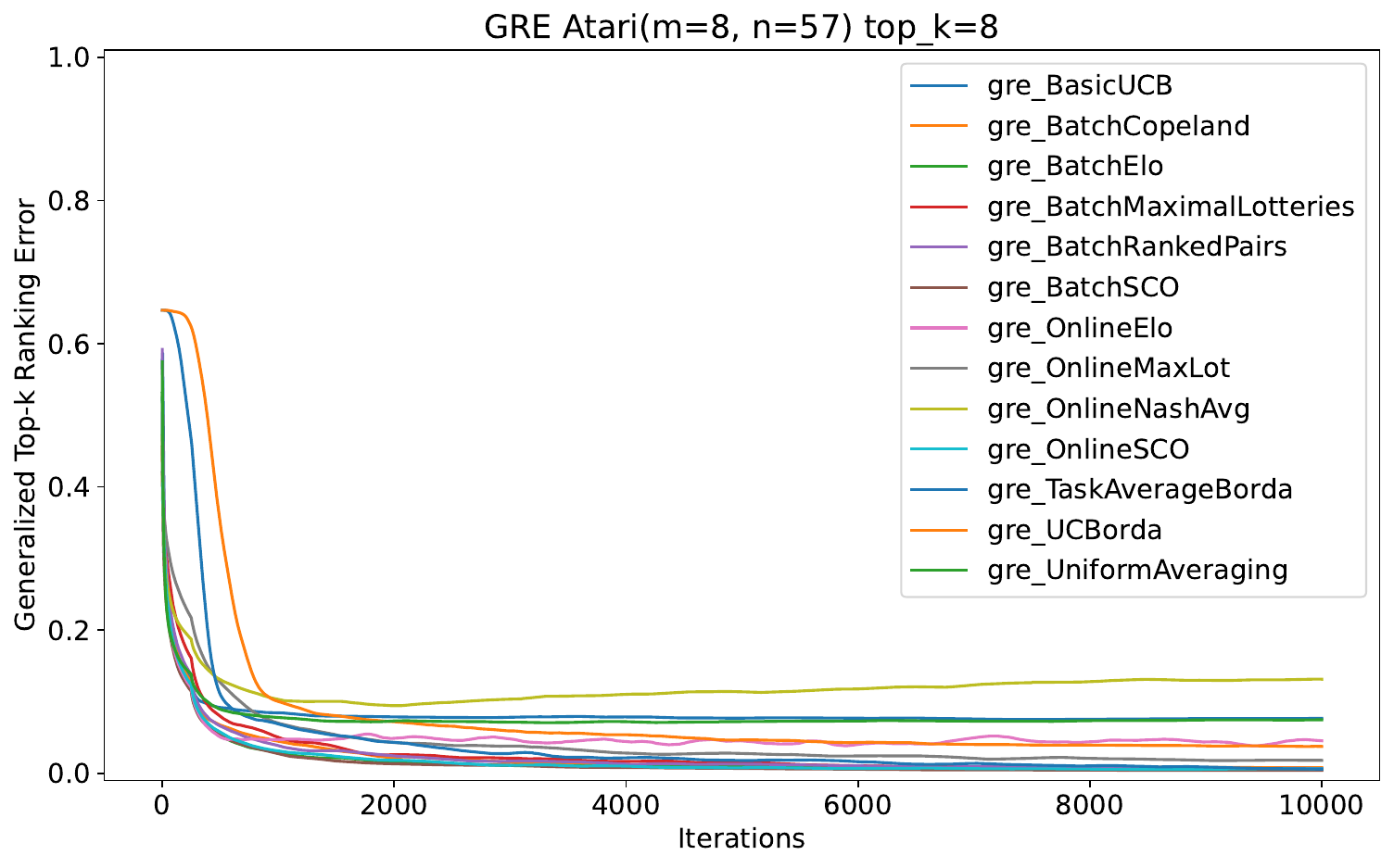} \\
\end{tabular}
\caption{Generalized Top-k Ranking Error on the Atari data set for all algorithms, for varying values of $k \in \{1, 2, 3, 8\}$.}
\label{fig:atari-gre-graphs}
\end{figure*}

Figure~\ref{fig:atari-gre-graphs} shows the graphs of GRE as a function of iterations for the Atari Agent57 data set, for various values of $k$.

\end{appendix}

\end{document}